\documentclass[lettersize,journal]{IEEEtran}
\usepackage{amsmath,amsfonts}
\usepackage{algorithmic}
\usepackage{array}
\usepackage[caption=false,font=normalsize,labelfont=sf,textfont=sf]{subfig}
\usepackage{textcomp}
\usepackage{stfloats}
\usepackage{url}
\usepackage{verbatim}
\usepackage{graphicx}
\usepackage{mathtools}
\usepackage{bm}
\usepackage{colortbl}
\usepackage{caption}
\usepackage{todonotes}
\usepackage{booktabs}
\usepackage{balance}
\usepackage{tikz}
\usepackage{scalefnt}
\usepackage[edges]{forest}
\usepackage{enumitem}
\tikzset{%
    parent/.style =          {align=center,text width=2.2cm,rounded corners=1pt, font=\small},
    child/.style =           {align=left,text width=1.9cm,rounded corners=3pt, font=\small},
    grandchild/.style =      {align=left,text width=9cm,rounded corners=3pt, font=\small},
    greatgrandchild/.style = {align=center,text width=1.5cm,rounded corners=3pt},
    referenceblock/.style =  {align=center,text width=1.5cm,rounded corners=2pt}
}

\newtheorem{definition}{Definition}
\newtheorem{example}{Example}

\newcommand{\bigo}[1]{\ensuremath{\mathcal{O}(#1)}}
\newcommand{\numc}{|\bm{N}_C|}
\newcommand{\numr}{|\bm{N}_R|}
\newcommand{\numi}{|\bm{N}_I|}
\newcommand{\deltai}{|\Delta^{\mathcal{I}}|}
\newcommand{\revision}[1]{{\color{black}#1}}
\newcommand{\fst}[1]{\textbf{#1}}
\newcommand{\snd}[1]{\underline{#1}}
\newcommand{\minrev}[1]{{\color{black}#1}}

\def\BibTeX{{\rm B\kern-.05em{\sc i\kern-.025em b}\kern-.08em
    T\kern-.1667em\lower.7ex\hbox{E}\kern-.125emX}}
\begin{document}
\title{Ontology Embedding: A Survey of Methods, Applications and Resources}
\author{Jiaoyan Chen, Olga Mashkova$^*$, Fernando Zhapa-Camacho$^*$, Robert Hoehndorf, Yuan He and Ian Horrocks
\thanks{Jiaoyan Chen is from University of Manchester \& University of Oxford in the UK. Fernado Zhapa-Camacho, Olga Mashkova and Robert Hoehndorf are from King Abdullah University of Science and Technology in Saudi Arabia. Yuan He and Ian Horrocks are from University of Oxford in the UK. $^*$ indicates Olga Mashkova and Fernando Zhapa-Camacho make equal contributions. More details of the author contributions are shown in the end.}
}
 
\markboth{Accepted by IEEE Transactions on Knowledge and Data Engineering (TKDE)}%
{Ontology Embedding: A Survey of Methods, Resources and Applications}

\maketitle

\begin{abstract}
Ontologies are widely used for representing domain knowledge and meta data, playing an increasingly important role in Information Systems, the Semantic Web, Bioinformatics and many other domains.
However, logical reasoning that ontologies can directly support are quite limited in learning, approximation and prediction.
One straightforward solution is to integrate statistical analysis and machine learning.
To this end, automatically learning vector representation for knowledge of an ontology i.e., \textit{ontology embedding} has been widely investigated.
Numerous papers have been published on ontology embedding, but a lack of systematic reviews hinders researchers from gaining a comprehensive understanding of this field.
To bridge this gap, we write this survey paper, which first introduces different kinds of semantics of ontologies and formally defines ontology embedding as well as its property of faithfulness.
Based on this, it systematically categorizes and analyses a relatively complete set of over 80 papers, according to the ontologies they aim at and their technical solutions including geometric modeling, sequence modeling and graph propagation.
This survey also introduces the applications of ontology embedding in ontology engineering, machine learning augmentation and life sciences, presents a new library mOWL and discusses the challenges and future directions.
\end{abstract}

\begin{IEEEkeywords}
Ontology, Ontology Embedding, Web Ontology Language, Knowledge Graph, Representation Learning 
\end{IEEEkeywords}

\section{Introduction}\label{sec:introduction}
Ontologies are formal, explicit and shared representations of
knowledge within a domain, with definitions and axioms for
concepts, properties, relations and other types of entities \cite{guarino2009ontology}.  
Ontologies have been a critical technology in Knowledge Management, Information Systems, the Semantic Web, Natural Language Processing and Artificial Intelligence, playing an increasingly important role in many fields such as Healthcare, Bioinformatics and E-commerce.  
A simple ontology can just be a set of concepts arranged in a hierarchy with the subsumption (inclusion) relationship between  concepts indicating that instances of one concept all belong to another.  
Such ontologies are capable of representing taxonomies of domains, such as the BBC Widelife Ontology \cite{raimond2010use} and the International Classification of Diseases (ICD) \cite{harrison2021icd}, dating back to Porphyrian Tree for presenting Aristotle's categories in the third century AD.  
Meanwhile, many data and knowledge management systems such as e-commerce platforms \cite{dong2018challenges} also adopt such simple ontologies for type information of data.

With the fast development of the Web in 1990s, representing and exchanging
data and knowledge on the Web became desirable. To this end, several
standards were proposed for defining more complex ontologies for
constructing the Semantic Web \cite{horrocks2008ontologies}\cite{hendler2001semantic}.  
In 1999, Resource Description Framework (RDF)\footnote{\url{https://www.w3.org/TR/rdf11-concepts/}} which
defines the syntax of triple \textit{(Subject, Predicate, Object)}
was proposed for representing data, and in 2000, the
vocabulary of RDF Schema (RDFS)\footnote{\url{https://www.w3.org/TR/rdf-schema/}}, was proposed for building ontologies as the schemas of data \cite{mcbride2004resource}.  
The vocabulary of RDFS can represent not only hierarchical concepts, but also instance membership to concepts, property hierarchies, and property domains and ranges.

The Web Ontology Language (OWL), including its second version OWL 2, was published upon the foundation of RDF and RDFS for building
ontologies that can represent more complex knowledge with logics such as the disjunction, conjunction and disjointness of concepts, and the existential and universal rules\footnote{\url{https://www.w3.org/TR/owl-features/}}. OWL was underpinned by
Description Logic --- a fragment of first-order logic with decidable
reasoning and efficient decision procedures \cite{baader2017introduction}. 
Many widely used ontologies, such as the Gene Ontology (GO) \cite{gene2019gene}, the Food Ontology (FoodOn) \cite{dooley2018foodon}, the DBpedia ontology \cite{auer2007dbpedia} and the aforementioned BBC Widelife Ontology, adopt OWL \cite{garcia2011analysis}.

Besides the formally and explicitly defined semantics, most real-world ontologies are also equipped with literals, including attribute values of instances defined by data properties, 
and meta-data associated with entities\footnote{In the community of ontology, a concept is modeled as a class, an instance corresponds to an individual in Description Logic, and the term ``entity'' includes class, instance and property. In the community of knowledge graph, the mention of an entity is actually equivalent to an instance. For clarity, we adopt the terms of the ontology community in this paper.} 
defined by annotation properties\footnote{\url{https://www.w3.org/TR/owl-ref/\#Annotations}}, representing information of name, definition, comment, image, version and so on.
For example, the concept \textit{obo:FOODON\_00002873} in FoodOn is associated with an English name ``okara'', a synonym ``soy pulp'', a long definition ``Okara, soy pulp, or tofu dregs is a pulp consisting of insoluble parts of the soybean that ...'', the source of this definition, and an image of okara \cite{dooley2018foodon}.  
These annotations, originally created for human understanding, contain important information that is often
complementary to the formal semantics. However, they cannot be utilised by symbolic reasoning.

In early 2010s, word embedding algorithms like Word2Vec were proposed to represent natural language words as low dimensional vectors, \revision{capturing their semantic relationships like correlations within the vector space \cite{mikolov2013efficient}.}
Similar ideas of representation learning were also applied to knowledge graphs (KGs) that are composed of relational facts, giving rise to some classic algorithms like TransE \cite{bordes2013translating} and RDF2Vec \cite{ristoski2016rdf2vec}. 
The instances and relations (i.e., object properties) are embedded with their semantics indicated by facts retained in the vector space. 
Take TransE as an example; the embedding of a KG, denoted as a mapping function $\bm{v}(\cdot)$ from
its instances and relations to vectors, is learned such that each relational fact
$(h, r, t)$ where the head instance $h$, the relation $r$ and the
tail instance $t$ correspond to the subject, predicate and object in
an RDF triple, respectively, is kept in the vector space as $\bm{v}(h) + \bm{v}(r) \approx \bm{v}(t)$.  Such embeddings not only enable \revision{machine learning and statistical algorithms to utilise}
the knowledge, but also
support neural-symbolic reasoning within a KG, with both approximation and prediction \cite{wang2017knowledge}\cite{chen2020review}.

Similarly, representation learning has also been applied to ontologies for embedding. 
Some early works such as \cite{nickel2017poincare} and \cite{vilnis2018probabilistic} proposed to embed triples with relations of transitivity, including those from WordNet \cite{miller1995wordnet}.
Their methods are applicable to simple ontologies with concept hierarchies. 
However, the semantics of RDFS and OWL ontologies are much more complex, and, accordingly, more advanced vector
representation models, which use high dimensional balls, boxes and axis-aligned cones for representing concepts, were recently proposed, following some early works including EmbedS \cite{diaz2018embeds} for RDFS ontologies, ELEmbeddings \cite{kulmanov2019embeddings} and E2R \cite{garg2019quantum} for OWL ontologies of Description Logic $\mathcal{EL}^{++}$ and $\mathcal{ALC}$, respectively.
Meanwhile, several complex embedding frameworks such as OPA2Vec \cite{smaili2019opa2vec} and OWL2Vec* \cite{chen2021owl2vec} were also proposed to embed both formal semantics and textual literals, often upon sequence learning methods.

Such ontology embedding methods have brought new solutions for ontology engineering and dramatically extended the application of ontologies. Many of them have been
verified in tasks with real-world data, including
concept subsumption inference \cite{chen2021owl2vec}, the domain application of protein--protein
interaction prediction \cite{smaili2019opa2vec}, and ontology augmented zero-shot and few-shot learning \cite{geng2021ontozsl}\cite{kulmanov2022deepgozero}.  
\revision{Besides ontology embedding, there are some other directions that attempt to combine ontology with machine learning or statistical approach, including fuzzy ontology standards such as Fuzzy OWL
\cite{stoilos2005fuzzy}, traditional ontology learning methods such as
the inductive logic programming system DL-Leaner
\cite{maedche2001ontology}\cite{lehmann2009dl}, and some neural-symbolic frameworks such as Logic Tensor Network \cite{badreddine2022logic}. 
In comparison with them,} 
ontology embedding methods focus on the automatic learning of the given knowledge's vector representations so as to supporting the integration with different machine learning and statistical models.

Briefly, there are quite a few results about ontology embeddings,
covering theoretic analysis, new methods and applications.
Although some survey papers for KG embedding involve ontologies
\cite{zhang2022knowledge}\cite{alam2023towards}\cite{xiong2023geometric}, but they only analyse those embedding methods that aim at relational facts, regarding simple ontologies as a kind of additional constraints.  
There is a shortage of systematic review to papers on ontology embeddings.  
Kulmanov et al. \cite{kulmanov2021semantic} reviewed some works in 2021 on using
machine learning for analyzing semantic similarity with ontologies.
Several ontology embedding methods including Onto2Vec
\cite{smaili2018onto2vec}, OPA2Vec \cite{smaili2019opa2vec} and EL
Embeddings \cite{kulmanov2019embeddings} are covered, but they are far from complete, especially considering there are quite a few  papers published after 2021.  

This survey aims to bridge the above gap, with
\textit{(i)} systematic categorization and comparison of the ontology embedding
methods according to the employed techniques and
the target ontologies,
\textit{(ii)} review of the applications in
supporting knowledge engineering, machine learning and life science
knowledge discovery together with benchmarks and metrics, 
\textit{(iii)} introduction and result demonstration of a library named mOWL \cite{mowl} that can support the implementation of ontology embedding methods, and \textit{(iv)} discussion on the challenges and future directions.  
\revision{This survey} has reviewed over 80 papers (around 40 of them are for new embedding methods) published in conferences and journals of Computer Science, AI and Bioinformatics, covering all the relevant works on ontology embedding, to the best of our knowledge. 
We believe it will benefit all the researchers who are interested in some topics among ontology, KG, knowledge representation, semantic embedding, semantic techniques, knowledge engineering, neural-symbolic integration, bioinformatics, and AI for life sciences.

The remainder of this paper is organized as follows. Section
\ref{sec:background} gives the background of ontology and semantic embedding.
\revision{
Section \ref{sec:method} reviews ontology embedding methods.  Sections \ref{sec:application} and \ref{sec:application2} review the applications.}
Section \ref{sec:systems} demonstrates mOWL. Section \ref{sec:challenge} presents our perspectives on challenges and
future directions. Section \ref{sec:conclusion} concludes the
paper.

\section{Background}\label{sec:background}

\subsection{Symbolic Knowledge Representation with Ontologies}
\subsubsection{Knowledge Graph (KG)}
In this paper, we distinguish KGs from ontologies.  We refer to KGs as those knowledge bases mainly composed of relational facts in RDF, following the definition in most KG embedding papers \cite{hogan2021knowledge}\cite{wang2017knowledge}.  
A KG can be denoted as
$\mathcal{G} = \left\{ \bm{I}, \bm{R}, \bm{T}\right\}$. $\bm{I}$
denotes a set of instances (also known as entities), corresponding individuals in Description Logic.
$\bm{R}$ denotes a set of binary relations.
$\bm{T}$ denotes a set of relational facts, i.e.,
$\bm{T} = \left\{(h, r, t) | h, t \in \bm{I}, r \in \bm{R} \right\}$, where $(h, r, t)$ is an RDF triple, $h$ and $t$, as the subject and object, are also called as the triple's head and tail, respectively.
One simple example is
$(\textit{Bob}, \textit{hasFather}, \textit{Alex})$.
Sometimes $(h, r, t)$ is also denoted in form of a relation $r(h, t)$.
For real-world KGs of the Semantic Web, each instance or relation should
be uniquely identified, usually by an Internationalized Resource
Identifier (IRI).

\subsubsection{RDF Schema (RDFS)}
RDFS can define either an ontological schema for a KG or an
independent ontology with the following main features.
\begin{itemize}[leftmargin=*]
\item RDFS can define a set of concepts (classes) $\bm{C}$ and assert the membership of instances using the built-in predicate
  \textit{rdfs:type}. For example, the triple
  $(\textit{Alex}, \textit{rdfs:type}, \textit{Father})$ represents
  that $Alex$ is an instance of \textit{Father}.
\item RDFS can define the subsumption between concepts with the
  built-in predicate \textit{rdfs:subClassOf}. Considering two concepts \textit{Father} and \textit{Parent} defined in $\bm{C}$. The triple
  $(\textit{Father}, \textit{rdfs:subClsasOf}, \textit{Parent})$
  represents that \textit{Parent} subsumes \textit{Father} and
  \textit{Father} is a sub-concept of \textit{Parent}.
\item RDFS can define the domain and range of a relation (i.e., object
  property) with the built-in predicates of \textit{rdfs:domain} and
  \textit{rdfs:range}, respectively. If the domain (resp. range) of a
  relation $r$ is a concept $C$, the heads (resp. tails) that are
  associated with $r$ must be declared or inferred to be 
  instances of $C$. For example, we can define the range of
  \textit{hasFather} to \textit{Father}, \textit{Parent} and/or
  \textit{Male}.  RDFS can also define the range of a data property
  with built-in data types.
\item RDFS can define the subsumption between two properties with the
  built-in predicate \textit{rdfs:subPropertyOf}, indicating that the
  instance pairs associated to one property all belong to those
  associated to another property. One example is
  $(\textit{hasFather}, \textit{rdfs:subPropertyOf}, \textit{hasParent})$.
\end{itemize}

\subsubsection{Description Logic (DL) and Web Ontology Language (OWL)}\label{sec:dl_owl}
OWL has different sub-languages and comes in multiple versions. The
complete languages, OWL Full and OWL 2 Full, are not decidable.  In
this part we mainly introduce \textit{(i)} the vocabularies of OWL 2 DL which are 
defined based on the DL fragment $\mathcal{SROIQ}$, and \textit{(ii)} some of its widely used sub-languages that are developed for different
scenarios with a better balance between knowledge expressivity and
reasoning complexity.  A signature consists of three finite sets of
symbols: a set $\bm{N}_I$ of individual names, a set $\bm{N}_C$ of
concept names, and a set $\bm{N}_R$ of role names. DL
$\mathcal{SROIQ}$ allows recursive concept definition as
\begin{align*}
  &\top\;|\; \bot\;|\; A\;|\; C\sqcap D\;|\;  C\sqcup D\;|\; \neg C\;|\; \exists r.C\;|\; \forall r.C\;|\; \\
  & \geq n r.C\;|\; \leq n r.C\;|\; \exists r.\textit{Self}\;|\;  \{a\}\nonumber
\end{align*}
where $\top$ is the top concept, $\bot$ is the bottom concept,
$A \in \bm{N}_C$ is an atomic (or named) concept, $r \in \bm{N}_R$ is
an atomic role (equivalent to a binary relation), $a \in \bm{N}_I$ is an individual, $C$ and $D$ are
themselves (possibly complex) concepts, \revision{$n$ is a number of
cardinality,} \minrev{ $\exists r.\textit{Self}$ is a concept that indicates the set of entities in the domain that are related by $r$ with themselves \cite{baader2017introduction}.} 
We say a concept is \textit{complex}
when it is constructed with one or multiple logical operators such as
$\sqcap$, $\sqcup$, $\exists$, $\forall$ and $\neg$.  A DL ontology
$\mathcal{O}$ can be composed of a TBox $\mathcal{T}$ and an ABox
$\mathcal{A}$.  The TBox defines logical background knowledge in the
form of concept subsumption axioms $C \sqsubseteq D$ (Generalized
Concept Inclusion, GCI), and role axioms for logical background
knowledge of role composition, role subsumption, and role
characteristics like functionality, transitivity and so on. 
Sometimes these role axioms are separately divided into an RBox, and accordingly the DL ontology is composed of a TBox, an RBox and an ABox.  
The ABox contains concrete data including concept assertions in form of $C(a)$, and role assertions in form of $r(a, b)$.  
With the defined logic, symbolic reasoners can be applied to infer hidden knowledge (i.e., entailment reasoning), check the ontology consistency and find justification that leads to inconsistency.

$\mathcal{ALC}$ and $\mathcal{EL}^{++}$ are two important fragments of DL that are widely investigated in ontology embedding.
$\mathcal{ALC}$ is known as Attributive Concept Language with
Complements and allows recursive concept definition with
$\top\;|\; \bot\;|\; A\;|\; C\sqcap D\;|\; C\sqcup D\;|\; \neg C\;|\;
\exists r.C\;|\; \forall r.C$ \cite{baader2017introduction};
$\mathcal{ALC}$ is a prototypical DL mainly studied because it has most of the major features.

The DL fragment $\mathcal{EL}^{++}$ which corresponds to OWL 2 EL profile allows recursive concept definition with $\top\;|\; \bot\;|\; A\;|\; C\sqcap D\;|\; \exists r.C\;|\; \{a\} $ \cite{baader2005pushing}.  
Due to a high knowledge representation capability but a polynomial time complexity in reasoning, DL $\mathcal{EL}^{++}$ is widely used by many real-world ontologies such as SNOMED CT~\cite{donnelly2006snomed}.
Example~\ref{ex:family} presents a toy family ontology of
DL $\mathcal{EL}^{++}$, with a TBox and an ABox.
\begin{example}\label{ex:family}
  The following $\mathcal{EL}^{++}$  ontology models a simple family domain:
  \begin{align*}
    \mathcal{T} = \{&\textit{Father}\sqsubseteq \textit{Parent}\sqcap \textit{Male},\;
                      \textit{Mother}\sqsubseteq \textit{Parent}\sqcap \textit{Female}, \\
                    &\textit{Child} \sqsubseteq \exists \textit{hasParent}.\textit{Father}, 
                      \textit{Child}\sqsubseteq\exists \textit{hasParent}.\textit{Mother}, \\
                      &\textit{hasParent}\sqsubseteq\textit{relatedTo}
                      \} \\
    \mathcal{A} = \{&\textit{Father}(\textit{Alex}), \textit{Child}(\textit{Bob}), \textit{hasParent}(\textit{Bob}, \textit{Alex})\}
  \end{align*}
\end{example}

\begin{figure}[!ht]
  \centering
  \includegraphics[scale=0.38]{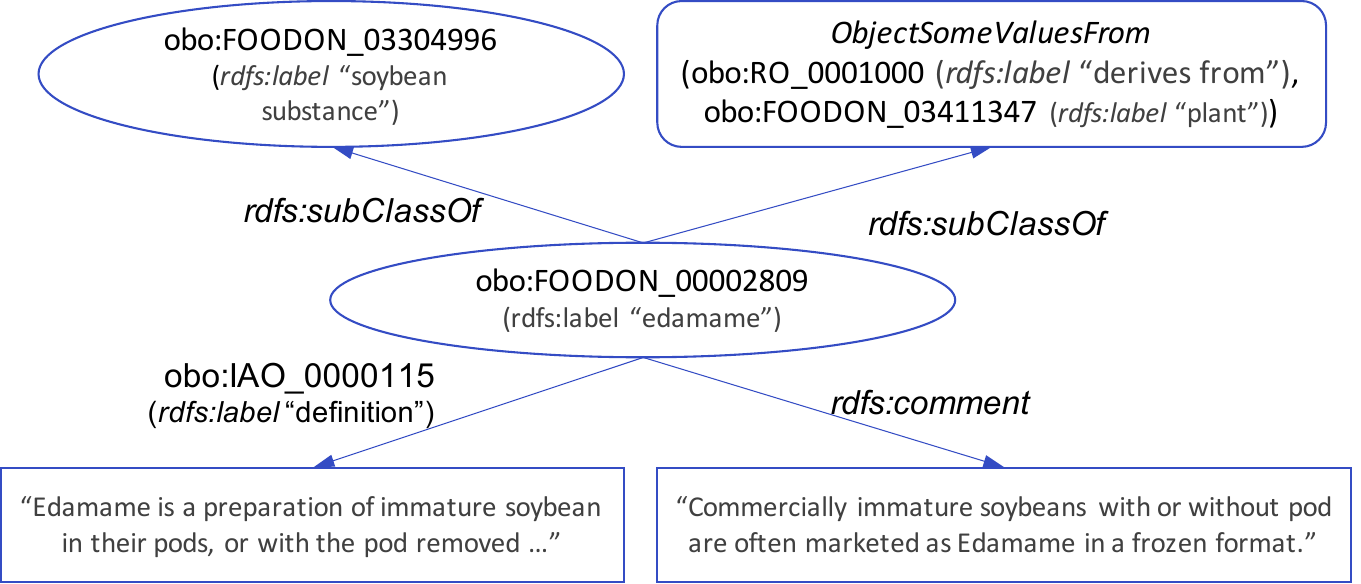}
  \caption{A fragment from the OWL ontology FoodOn \cite{dooley2018foodon}.}
  \label{fig:foodon}
\end{figure}

The ontology in Example~\ref{ex:family} can be implemented with the vocabularies defined in the standards of
RDF, RDFS and OWL; for example, 
\textit{rdfs:subClassOf} for
concept subsumption, \textit{owl:ObjectSomeValuesFrom} for the existential quantification $\exists r.C$ and
\textit{owl:ObjectAllValuesFrom} for the universal quantification $\forall r.C$.  
Figure~\ref{fig:foodon} presents a fragment for the concept  \textit{obo:FoodON\_00002809} (``edamame'') which is a sub-concept of a named concept \textit{obo:FOODON\_03304996} (``soybean substance'') and an existential quantification with the property of \textit{obo:RO\_0001000} (``derives from'') and the concept of \textit{obo:FOODON\_03411347} (``plant'').

\subsubsection{Ontology Literals}
The literals of an ontology are mainly defined in two approaches.  (1)
The first approach is to associate instances with literals by datatype
properties, which are sometimes known as attributes
and whose values can be of different types such as natural language
phrases and long text, real values, data and time, category, image and
domain specific sequence (e.g., gene sequence). 
For example, instances of \textit{Person} may have address, height, birth data and so on. 
(2) The second approach is to associate entities with meta information by annotation properties.
In real-world ontologies, most such literals are in form of text.  
For example, in the ontology fragment in Figure \ref{fig:foodon}, the concepts and properties are annotated with names by the built-in vocabulary of
\textit{rdfs:label}\footnote{The textual value often has a language
  tag and is regarded as English by default. The entity IRIs sometimes
  also indicate the name information.}, and
\textit{obo:FOODON\_00002809} is annotated with two long sentences ---
a comment by \textit{rdfs:comment} and a definition by the ad-hoc
annotation property \textit{obo:IAO\_0000115}.  Some other literals
such as links to external sources, editors and images are also widely used.
Although all these literals include important information, as informal semantics, they are often ambiguous and cannot be used for inference by symbolic ontology reasoners.
\textit{In ontology embedding, ontologies either with or without
  literals are considered.}

\subsubsection{Target Ontologies}
In our context of ontology embedding, ``ontology'' refers either to a TBox, or any DL ontology with a non-empty TBox, i.e., $(TBox, ABox, RBox)$ or $(TBox, ABox)$ with $TBox \not= \emptyset$.
Different embedding methods aim at different semantics of an ontology.
Accordingly, we divide the ontologies considered in the current ontology embedding works into four kinds: 
\begin{itemize}[leftmargin=*]
\minrev{
\item \textbf{Simple Ontology} which refers to DL ontologies that have only a TBox composed of the top concept $\top$ (e.g., \textit{owl:Thing} in OWL ontologies), named concepts and subsumption axioms between named concepts. They are equivalent to taxonomies composed of hierarchical classes.
\item \textbf{Complex Ontology} which refers to DL ontologies that have a TBox containing any concept definitions of DL
$\mathcal{SROIQ}$ beyond $\top$. These ontologies may also contain an ABox, an RBox, or both.
\item \textbf{Ontology with Literals} which refers to simple or complex ontologies that have literals.
\item \textbf{Ontology with KG} which refers to KGs composed of large scale relational facts, equipped with DL ontologies that has a TBox (and an RBox in option) as the KG's schema.}
\end{itemize}

\subsection{Semantic Embedding}

In this part, we first summarise word embedding and KG embedding,  and then give formal definitions and properties of \minrev{general embedding and} ontology embedding.

\subsubsection{Word Embedding} 
Word embedding algorithms such as Word2Vec \cite{mikolov2013efficient}
and GloVe \cite{pennington2014glove} learn vector representations of
tokens (which are either words or sub-words) from a large corpus
concerning their semantics such as co-occurrence in the sentences.
Taking Word2Vec as an example, it learns a Feed-Forward Neural Network
from natural language sentences by one of the two auto-encoding
architectures --- continuous skip-gram which predicts the surrounding
tokens of each token and continuous Bag-of-Words which predicts
a token based on its surrounding tokens.  For each token, the hidden
layer output of the network is its embedding.
Tokens with more similar meanings are expected to have higher vector similarities.
Such embedded semantics can partially support analogical reasoning, e.g., $\bm{v}(king) - \bm{v}(father) \approx \bm{v}(queen) - \bm{v}(mother)$.

Embeddings by Word2Vec and GloVe are \textit{non-contextual},
which means each token has one unique vector representation no matter what surrounding tokens it has.
Recently, with the development of Transformer-based encoder architectures \cite{vaswani2017attention} and Pre-trained Language
Models (PLMs) like BERT \cite{kenton2019bert}, \textit{contextual word embeddings} have been widely developed and adopted.  
Taking BERT as an example, it learns a
Transformer architecture from a corpus by predicting the masked token
in each sentence and the next sentence of a given sentence.  The
vector of a token is based on the attentions
from itself and its surrounding tokens in the sentence.
Given a sentence ``the bank robber was seen on the river bank'', the first ``bank'' and the second ``bank'' have different vectors due to different surrounding tokens.  
Such contextual word embeddings encode more semantics and often perform better than non-contextual word embeddings in many tasks.

Besides natural language text, the above contextual and non-contextual
word embedding techniques are also applicable to other kinds of
sequential data, such as BioVec for biological sequences like genes
and proteins \cite{asgari2015continuous}, and Node2Vec for paths from graphs \cite{grover2016node2vec}.  
Their great success in many domains also motivate researchers to applying them to KGs and ontologies.  For simplicity, in this paper
\textit{we call word embedding as well as representation learning for other sequential data as \textbf{sequence learning}}.

\subsubsection{Knowledge Graph Embedding}
These methods learn vector representations of instances and relations
from the relational facts in a KG with their semantics retained in the
vector space \cite{wang2017knowledge}.  In general, typical technical
solutions can be divided into 
\textit{geometric modeling} such as the translational method TransE
\cite{bordes2013translating}, \textit{tensor decomposition} such as
the bilinear method DistMult \cite{yang2015embedding}, \textit{neural
  network modeling} such as ConvE \cite{dettmers2018convolutional}, \textit{random
  walk-based sequence modeling} such as RDF2Vec
\cite{ristoski2016rdf2vec}.  TransE has been introduced
in Section \ref{sec:introduction}.  Taking RDF2Vec as another example,
it first conducts random walk for
extracting paths composed of instances and
relations, and then learns a Word2Vec model
for encoding them. 
KG embedding has also been extended to
encode semantics beyond relational facts, especially textual literals in combination with word embedding methods
\cite{gesese2021survey}\cite{pan2023large}, and logics such as horn rules,
schemas and constraints with additional modeling methods
\cite{zhang2022knowledge}\cite{xiong2023geometric}.

\subsubsection{Formal Definitions of Embedding}\label{sec:fde}

\minrev{Although the term ``embedding'' has been widely used in contexts of machine learning, NLP and KG, we find it useful to explicitly make our understanding of embedding clear here as it will provide a guide for analyzing and classifying ontology embedding methods.
We start with the definition of ``embedding'' as it is used in mathematics:}
\begin{definition}[Embedding (mathematics)]
  An {\em embedding} is an injective and structure-preserving map between two
  mathematical structures which can be algebraic, topological, or
  geometrical structures.
\end{definition}
In machine learning, embedding is used in a somewhat different sense:
\revision{
\begin{definition}[Embedding (machine learning)]
  An {\em embedding} $e$ is a learned mapping between the elements
  of a mathematical structure and the elements of some structure $\mathcal{S}$.
\end{definition}
}
The notion of ``embedding'' in machine learning is therefore related
to, but less strict than, a structure-preserving map, and embeddings
are representations learned from data, largely built upon the
foundation of representation learning \cite{bengio2013representation}.
\minrev{The learned representations are usually not
arbitrary but rather aim to preserve some ``semantics'' of the
original data. We capture this by defining the notion of
``faithfulness'':
\begin{definition}[Faithfulness of embedding (machine learning)]
  An {\em embedding} (machine learning) $e$ is ``faithful'' if $e$ converges to some embedding (mathematics) $f$.
\end{definition}

\subsubsection{Definitions and Properties of Ontology Embedding}\label{sec:dpoe}

An ontology has a syntactic structure which consists of a set of logical symbols like connectives and quantifiers, a set of non-logical symbols including constants, functions and relations, a set of well-formed formulas constructed from the logical and non-logical symbols, a set of axioms (a subset of the well-formed formulas), and a set of inference rules for deriving new formulas from the axioms and previously derived formulas. 
The signature of the ontology includes concept, individual and relation symbols.
The ontology's model structure over its signature consists of a non-empty set $D$, called the domain or universe of the model, an interpretation function $I$ that assigns each individual symbol to an element of $D$, each concept symbol to a subset of $D$, and each binary relation symbol to a binary relation on $D$ (i.e., a subset of $D^2$).
}
The semantic structure of the ontology $\mathcal{O}$, denoted as $Mod(\mathcal{O})$, consists of the collection of all model structures that satisfy the axioms in $\mathcal{O}$. $Mod(\mathcal{O})$ can be seen as a class or a category \revision{in the sense of category theory}, depending on the context. 
%
\minrev{With these background, we formally define ontology embedding and its faithfulness:
\begin{definition}[Ontology embedding]
  Let $\mathcal{O}$ be an ontology with signature
  $\Sigma_\mathcal{O}$. An ontology embedding is an embedding (machine learning) $e$ of the
  term algebra $T(\Sigma_\mathcal{O})$ generated by the signature
  $\Sigma_\mathcal{O}$ and the well-formedness
  rules of the underlying
  Description Logic for constructing concept descriptions, role
  expressions, and axioms.
\end{definition}
%
\begin{definition}[Faithful ontology embedding]
  An embedding of an ontology $\mathcal{O}$ is {\em faithful} if it converges to an embedding (mathematics) which preserves some mathematical structure $\mathcal{S}$ of $\mathcal{O}$.
\end{definition}
Here, we only consider embeddings for specific ontologies (with a
specific signature), not embeddings of an entire logic.
The notion of ``faithfulness'' for embeddings requires specifying a
mathematical structure $\mathcal{S}$ to which the embeddings are
faithful. 
There are at least three mathematical structures that can be assigned to ontologies:
  (1) the syntactic structure, potentially combined with
  syntactic inference rules (i.e., the deductive calculus);
  (2) a single arbitrary model of the ontology, i.e., a
  model structure in which all ontology axioms are true; 
and (3) the semantic structure of the ontology~\cite{dlhandbook}.
The mathematical structure that the embedding aims to preserve can also be used to classify and distinguish ontology embedding models.
}


\revision{
Another property of ontology embedding is ``interpretability''. In machine learning, there is no universally-agreed definition of interpretability~\cite{linardatos2020explainable};
in \cite{doshi2017towards}, interpretability is described as ``the ability to explain or to present in understandable terms to a human''.
In the context of ontology embedding, we can strictly define ``interpretability'' as the ability to reconstruct both symbols and composition rules from their embeddings. Faithful ontology embeddings are interpretable in that sense since injectivity ensures that each symbol can be uniquely recovered from its embedded representation, and the preservation of a mathematical structure of an ontology inherent in faithful embeddings allows for the restoration of the composition rules between symbols.
\minrev{A non-faithful ontology embedding does not support full
  restoration, but may still allow restoring a part of the symbolic
  semantics and justifying the inference with the embeddings. This
  characteristic is important for assessing ontology embedding methods
  since it reflects to what extent humans, who inherently engage in
  symbolic reasoning, can read off ontology axioms and understand the
  reasoning from the constructed embedding.}
}

\section{Ontology Embedding Methods}\label{sec:method}

In this section, we first analyse the main technical solutions that
have been commonly adopted for ontology embedding
(Section~\ref{sec:mc}), and then review the ontology embedding works
for each kind of ontologies (Section \ref{sec:soe} to \ref{sec:oekg}).

\subsection{General Technical Solutions}\label{sec:mc}
\begin{figure}[!ht]
     \centering
         \includegraphics[width=0.48\textwidth]{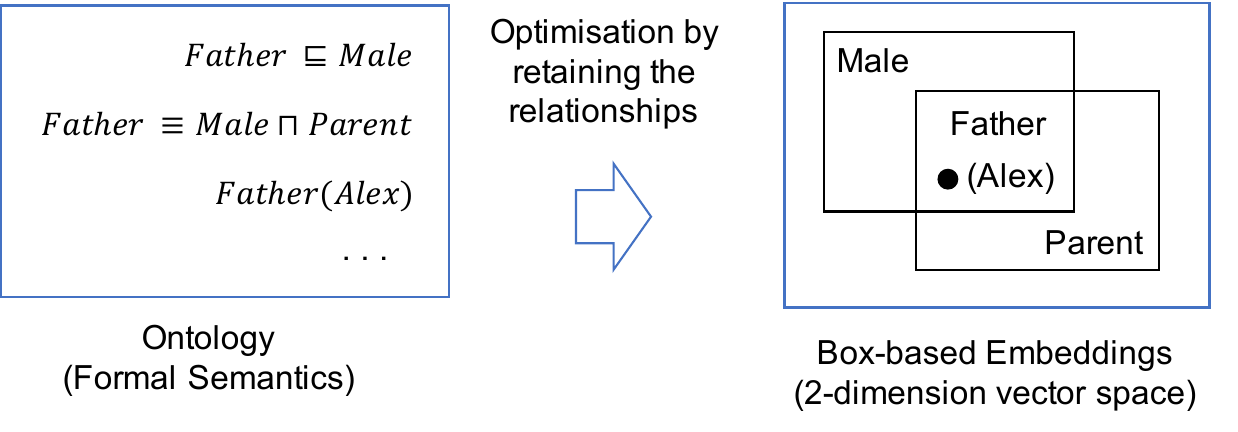}
         \vspace{-0.2cm}
         \caption{Demonstration of the ontology embedding solution of geometric modeling with the examples of ELBE and $\text{Box}^\text{2}\text{EL}$.}
         \label{fig:gm}
\end{figure}

\begin{figure*}[!ht]
     \centering
         \includegraphics[width=0.7\textwidth]{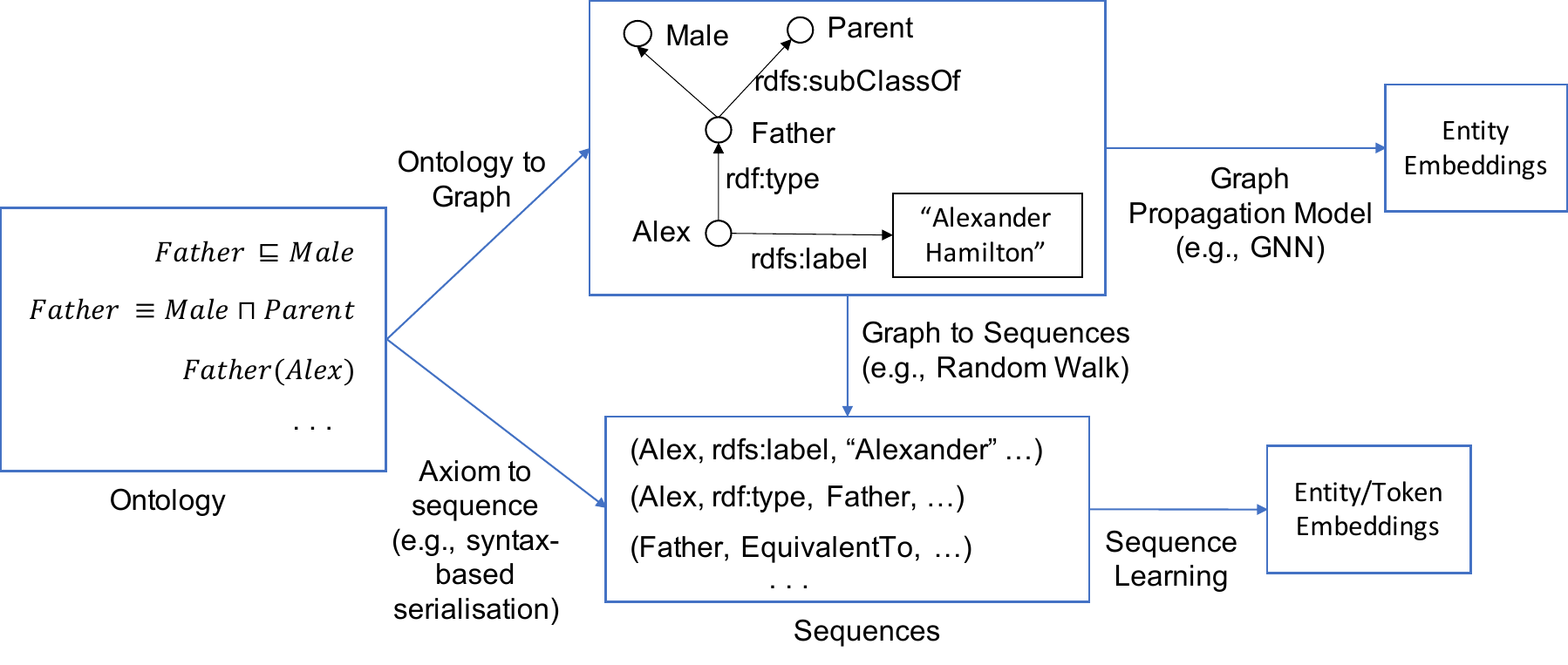}
         \vspace{-0.2cm}
         \caption{Demonstration of the ontology embedding solutions of sequence modeling and graph propagation.}
         \label{fig:sl}
\end{figure*}

The technical solutions that are commonly adopted by the
current ontology embedding methods include the following:
\begin{itemize}[leftmargin=*]
\item \textbf{Geometric Modeling} aims to generate embeddings that are faithful to some model structures of an ontology; they generate (or approximate)
  logical models of an ontology by interpreting concepts as
  geometric regions, interpret individuals as members of these
  regions, and relations as pairs of points standing in some geometric relation.  
  Take the embedding algorithms ELBE
  \cite{peng2022description} and $\text{Box}^\text{2}\text{EL}$
  \cite{jackermeier2024dual} for the DL $\mathcal{EL}^{++}$ as an
  example.  They model an instance as a high dimensional point, i.e.,
  one single vector, and models a concept as a high dimensional
  axis-aligned box represented by one vector for the box center and
  one vector for the box offset.  The instantiation relation between
  an instance and a concept is modeled as the instance's point lying
  within the concept's box.  The basic idea is demonstrated in Figure
  \ref{fig:gm} with a toy example. 
  \revision{
  These types of embeddings are highly interpretable since they induce geometric relations  between geometric objects representing relations, concepts and individuals which align with ontology axioms. 
  }
\item \textbf{Sequence Modeling} first transforms the ontology axioms
  and literals into sequences composed of entities (and tokens), and then adopts a sequence learning model to learn their embeddings. The basic
  idea is demonstrated in Figure \ref{fig:sl}.  Take  $\text{OWL2Vec}^{*}$ \cite{chen2021owl2vec} as an
  example.  It first extracts sequences composed of entities and
  textual literals from an OWL ontology by mapping the ontology to a
  graph and performing multiple random walks over the graph, and then
  learns a word embedding model from the sequences for encoding the
  entities and tokens in the text.  These methods are usually
  partially faithful to correlations and co-occurrences of symbols in
  the axioms of the ontology\revision{, and thus they have low interpretability}.
\item \textbf{Graph Propagation} represents an ontology by a
  (multi-relation) graph with initial node representations, and then learns a graph propagation model for new node representations. The basic idea
  is also demonstrated in Figure~\ref{fig:sl}.  For example, for concept matching, the work
  \cite{hao2023ontology} uses two Graph Convolutional Networks to
  learn embeddings of two cross-ontology concepts, respectively, by
  propagating the initial word embeddings of their surrounding concepts.
  \revision{Methods of this type are commonly interpretable to a limited degree since they preserve mainly the concept hierarchy.}
\end{itemize}

Table \ref{table:ts} gives a summary of these technical solutions,
including an analysis of the pros and cons according to two dimensions
--- \textbf{faithfulness} with regard to different mathematical
structures of an ontology, and \textbf{interpretability} \revision{which indicates how transparency the reasoning in the vector space is.}
Table~\ref{table:ts} also classifies different ontology embedding methods of each technical solution based on the ontologies that they can deal with (i.e., the \textbf{targeted ontologies}).

\begin{table*}[t]
\footnotesize{
\centering
\renewcommand{\arraystretch}{1.3}
\begin{tabular}[t]{m{1.4cm}<{\centering}|m{3.3cm}<{\centering}|m{2.8cm}<{\centering}|p{8cm}  }\hline
 \textbf{Solution}  &\textbf{Pros} & \textbf{Cons} &\textbf{Targeted Ontologies and Citations}    \\ \hline
Geometric Modeling & High interpretability; high faithfulness to the semantic structure  & Hard to integrate literals; not support many features of OWL &\begin{tabular}{l}Simple Onto.: \cite{vendrov2015order}\cite{athiwaratkun2018hierarchical}\cite{vilnis2018probabilistic}\cite{li2018smoothing}\cite{patel2020representing}\cite{dasgupta2020improving}\cite{nickel2017poincare}
\cite{ganea2018hyperbolic}\cite{lu2019learning}\cite{li2021hypon}\cite{pan2021hyperbolic}
\\Complex Onto.: \cite{gutierrez2018from}\cite{kulmanov2019embeddings}\cite{garg2019quantum}\cite{ozccep2020cone}\cite{mondal2021emel++} \cite{peng2022description}\cite{xiong2022faithful}\cite{ozcep2023embedding}\cite{lacerda2023strong}\cite{zhapa2023cate}\cite{jackermeier2024dual} \\Onto. \& Lit.: \cite{geng2021ontozsl} \\ Onto. \& KG: \cite{diaz2018embeds}\cite{lv2018differentiating}\cite{hao2019universal}\cite{xiang2021ontoea}\cite{yu2023geometry}\cite{huang2023concept2box}\cite{wang2024embedding}\end{tabular}     \\ \hline
Sequence Modeling & Extensible to different ontologies; support literals; learn correlations as effective features for downstream tasks& Low interpretability; only partial faithful to correlations and co-occurrences  &    \begin{tabular}{l}
Complex Onto.: \cite{alshahrani2017neuro}\cite{smaili2018onto2vec}\cite{holter2019embedding}  \\ Onto. \& Lit.: \cite{xiang2015ersom}\cite{jayawardana2017deriving}\cite{kolyvakis2018deepalignment}\cite{smaili2019opa2vec}\cite{dong2019imposing}  \cite{liu2020concept} \cite{chen2021owl2vec}\cite{nguyen2021biomedical}\cite{liu2021self}\cite{zhang2022ontoprotein}\cite{chen2023contextual}\cite{gosselin2023sorbet}\cite{he2024language} \end{tabular}  \\ \hline
Graph Propagation & Well embed the graph especially the concept hierarchy  & Low interpretability; only partially faithful to the graph structure  & \begin{tabular}{l}Simple Onto.: \cite{ma2021hyperexpan}  
\\ Onto. \& Lit.: \cite{hao2021medto}\cite{hao2023ontology}  \end{tabular}    \\ \hline
\end{tabular}
\caption{\footnotesize Important characteristics (pros and cons) of the three technical solutions, and the ontologies that main methods of each technical solution aim at. Onto. and Lit. are short for ontology and literal, respectively.
}\label{table:ts}
}
\end{table*}

\subsection{Embedding Simple Ontology}\label{sec:soe} 

\begin{figure*}[!ht]
     \centering
\begin{forest}
    for tree={forked edges, grow'=0, draw, rounded corners, node options={align=center,}, text width=2.7cm,},
    [Embedding Simple Ontology, for tree={fill=white!45, parent, text width=2cm, font=\scriptsize}
        [Embedding Method, for tree={fill=white!45, child, text width=2cm, node options={align=center,},font=\scriptsize}       
            [Box embedding \cite{vilnis2018probabilistic}\cite{li2018smoothing}\cite{patel2020representing}\cite{dasgupta2020improving}, fill=white!30, grandchild,font=\scriptsize]  
            [Cone embedding \cite{vendrov2015order}\cite{athiwaratkun2018hierarchical}\cite{ganea2018hyperbolic}, fill=white!30, grandchild,font=\scriptsize]
            [Hyperbolic distance-based embedding \cite{nickel2017poincare}\cite{lu2019learning}\cite{pan2021hyperbolic}\cite{li2021hypon}, fill=white!30, grandchild,font=\scriptsize]   
            [Graph propagation \cite{ma2021hyperexpan}, fill=white!30, grandchild,font=\scriptsize]  
        ]
        [Embedding Space, for tree={fill=white!45, child, text width=2cm, node options={align=center,}, font=\scriptsize}
            [Euclidean space \cite{vendrov2015order}\cite{athiwaratkun2018hierarchical}\cite{vilnis2018probabilistic}\cite{li2018smoothing}\cite{patel2020representing}\cite{dasgupta2020improving}, fill=white!30, grandchild, font=\scriptsize]
            [Hyperbolic space \cite{nickel2017poincare}\cite{ganea2018hyperbolic}\cite{lu2019learning}\cite{pan2021hyperbolic}\cite{li2021hypon}\cite{ma2021hyperexpan}, fill=white!30, grandchild, font=\scriptsize]
        ]
    ]
\end{forest}
\caption{Dimensions, their values and corresponding works of embedding simple ontology.}
\label{fig:simpleonto}
\end{figure*}
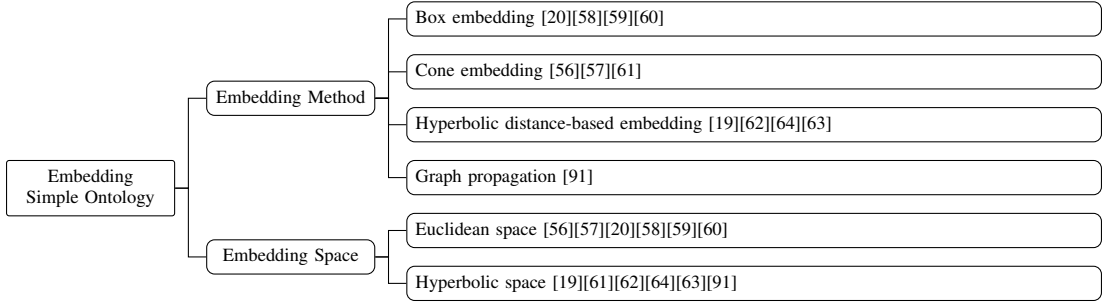

The concept hierarchies of an ontology imply its basic structural semantics. We refer to ontology embedding that solely
consider concept hierarchies as simple ontology embedding as it
omits more complex logical semantics. Figure \ref{fig:simpleonto} presents two dimensions --- Embedding Method and Embedding Space, their values and corresponding works of simple ontology embedding.

Geometric modeling is the most prevalent technique for simple ontology
embedding. This includes box embedding in Euclidean space, hyperbolic
distance-based embedding in hyperbolic space, cone embedding for both
geometries, and graph propagation in hyperbolic space.

In the context of Euclidean space, a key approach is to devise a
function that preserves the hierarchical order of entities. A typical
work by \cite{vendrov2015order} proposes using the reversed product
order, essentially forming a Euclidean cone where each entity's
embedding value is at least that of its parent entity. Building on
this, \cite{athiwaratkun2018hierarchical} extends to encode entities
using probability densities instead of points. The box embedding is
another typical construction which simulates the hierarchical ordering
with hyper-rectangles, where a child entity's box is consumed within
its parent entity's box. Unlike cone embeddings, which are typically
parameterized by their apex, box embeddings require two vectors
representing the minimal and maximal coordinates in the
hyper-rectangle. To further improve box embeddings,
\cite{li2018smoothing} explores a probabilistic relaxation to achieve
a smoother distribution, \cite{patel2020representing} investigates
encoding dual hierarchical relationships (hypernym and meronym) at the
same time, and \cite{dasgupta2020improving} addresses the local
identifiability issue, proposing the Gumbel-box process as a solution.

Hyperbolic space, with its expansive property and theoretical
underpinning, is particularly suitable for representing hierarchical
structures. The Poincar\'e embedding \cite{nickel2017poincare} is a typical approach that minimizes
hyperbolic distances between related entities while maximizing the
separation from unrelated ones in a unit Poincar\'e ball. This spatial
arrangement places more general entities near the origin and more
specific entities closer to the boundary, reflecting their
hierarchical depth. However, numerical instabilities near the boundary
of the manifold are a known challenge. To address this,
\cite{nickel2018learning} investigates an alternative hyperbolic
model, the Lorenz (or Hyperboloid) model, while
\cite{pan2021hyperbolic} proposes the extended Poincar\'e ball with
geometric distortion. To augment the embedded semantics,
\cite{lu2019learning} and \cite{li2021hypon} explores the integration
of Poincar\'e embeddings with pre-trained word embeddings, while
\cite{ma2021hyperexpan} utilises the hyperbolic graph convolutional
networks (HGCN) \cite{chami2019hyperbolic} for aggregating
neighbourhood information. In hyperbolic space, establishing geometric
shapes like boxes and cones, which are straightforward in Euclidean
space, requires more nuanced considerations. A notable contribution in
this line is the hyperbolic entailment cone by
\cite{ganea2018hyperbolic}, which not only preserves transitivity but
also enables direct prediction of entity subsumptions through its
construction.

The techniques discussed here aim to achieve high geometric
interpretability in encoding hierarchies, whether preserving order or
exploiting geometric properties. Although not all cited works
specifically test with ontology concept hierarchies, their methodologies
are readily applicable to such structures.

\subsection{Embedding Complex Ontology}\label{sec:coe}

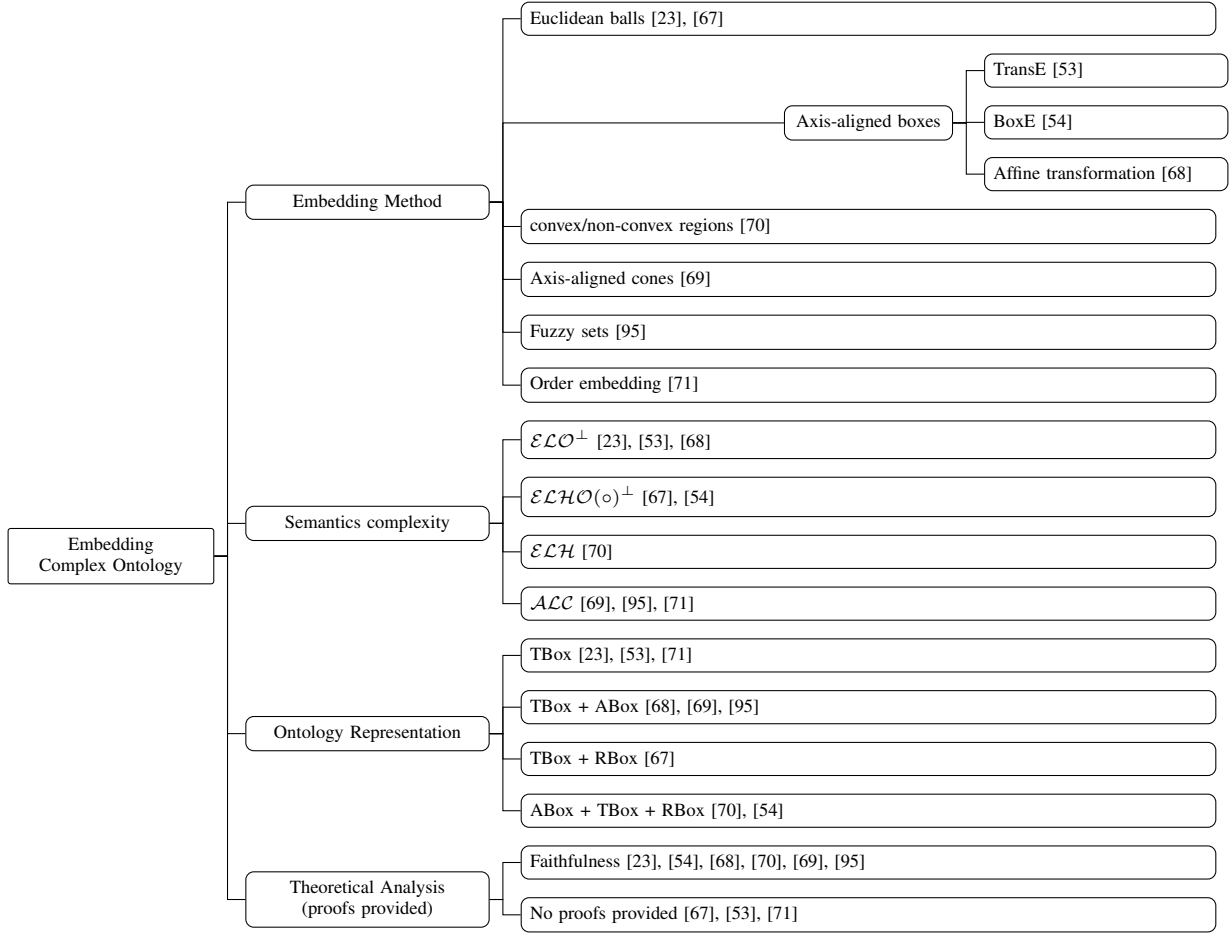
\begin{figure*}[!ht]
     \centering
\begin{forest}
    for tree={forked edges, grow'=0, draw, rounded corners, node options={align=center,}, text width=2.7cm,},
    [Embedding Complex Ontology, 
    for tree={fill=white!45, parent, text width=2.5cm, font=\scriptsize}
        [Embedding Method, for tree={fill=white!45, child, text width=3cm, node options={align=center,}, font=\scriptsize}
            [Euclidean balls~\cite{kulmanov2019embeddings,mondal2021emel++} , fill=white!30, grandchild, font=\scriptsize]
            [Axis-aligned boxes , for tree={fill=white!30, grandchild, text width=2cm, node options={align=center,},font=\scriptsize}
            [TransE~\cite{peng2022description} , fill=white!30, child, text width=3cm, font=\scriptsize]
            [BoxE~\cite{jackermeier2024dual} , fill=white!30, child, text width=3cm, font=\scriptsize]
            [Affine transformation~\cite{xiong2022faithful}  , fill=white!30, child, text width=3cm, font=\scriptsize]]
            [convex/non-convex regions~\cite{lacerda2023strong} , fill=white!30, grandchild,font=\scriptsize]
            [Axis-aligned cones~\cite{ozcep2023embedding} , fill=white!30, grandchild,font=\scriptsize]
            [Fuzzy sets~\cite{tang2022falcon} , fill=white!30, grandchild, font=\scriptsize]
            [Order embedding~\cite{zhapa2023cate} , fill=white!30, grandchild, font=\scriptsize]
        ]
        [Semantics complexity, for tree={fill=white!45, child, text width=3cm, node options={align=center,}, font=\scriptsize}
            [$\mathcal{ELO^{\bot}}$~\cite{kulmanov2019embeddings,peng2022description,xiong2022faithful} , fill=white!30, grandchild, font=\scriptsize]
            [$\mathcal{ELHO(\circ)^{\bot}}$~\cite{mondal2021emel++,jackermeier2024dual} , fill=white!30, grandchild, font=\scriptsize]
            [$\mathcal{ELH}$~\cite{lacerda2023strong} , fill=white!30, grandchild, font=\scriptsize]
            [$\mathcal{ALC}$~\cite{ozcep2023embedding,tang2022falcon,zhapa2023cate} , fill=white!30, grandchild, font=\scriptsize]
        ]
        [Ontology Representation, for tree={fill=white!45, child, text width=3cm, node options={align=center,}, font=\scriptsize}
            [TBox~\cite{kulmanov2019embeddings,peng2022description,zhapa2023cate} , fill=white!30, grandchild, font=\scriptsize]
            [TBox + ABox~\cite{xiong2022faithful,ozcep2023embedding,tang2022falcon} , fill=white!30, grandchild, font=\scriptsize]
            [TBox + RBox~\cite{mondal2021emel++} , fill=white!30, grandchild, font=\scriptsize]
            [ABox + TBox + RBox~\cite{lacerda2023strong,jackermeier2024dual} , fill=white!30, grandchild, font=\scriptsize]
        ]
        [Theoretical Analysis (proofs provided), for tree={fill=white!45, child, text width=3cm, node options={align=center,}, font=\scriptsize}
        [Faithfulness~\cite{kulmanov2019embeddings,jackermeier2024dual,xiong2022faithful,lacerda2023strong,ozcep2023embedding,tang2022falcon}, fill=white!30, grandchild, font=\scriptsize]
            [No proofs provided~\cite{mondal2021emel++,peng2022description,zhapa2023cate}, fill=white!30, grandchild, font=\scriptsize]
        ]
                    %
    ]
\end{forest}
\caption{Dimensions, their values and corresponding works of embedding complex ontology.}
\label{fig:ontocomplexdim}
\end{figure*}

Complex ontologies are mainly embedded by geometric
modeling.  The methods aim to map concepts, individuals and roles  into a continuous
vector space (e.g., $\mathbb{R}^n$) where they can be represented as
points or geometrical regions. This allows to capture some aspects of
semantics of the underlying ontology by means of geometric properties
of the embedding space.
We analyze complex ontology embedding methods from four perspectives: embedding method, semantics
complexity, strategy to embed ABox and RBox axioms, and theoretical
analysis; corresponding categorization and related works
are shown in Figure~\ref{fig:ontocomplexdim}.

Multiple geometric models have been developed for the
lightweight DL $\mathcal{EL}^{++}$.
ELEmbeddings~\cite{kulmanov2019embeddings} and
EmEL++~\cite{mondal2021emel++} represent named concepts as
n-dimensional Euclidean balls, which cannot faithfully model
concept intersection since the intersection of two balls is no longer a
ball in $\mathbb{R}^n$.  To address this issue, boxes have been adopted~\cite{peng2022description}\cite{xiong2022faithful}\cite{jackermeier2024dual}.
These methods can be categorized further by the relation model implemented
within their frameworks: ELBE~\cite{peng2022description} utilizes
TransE~\cite{bordes2013translating}; this model based on vector
translations cannot faithfully represent 1-to-N, N-to-N and N-to-1
relations.  To overcome this limitation,
$\text{Box}^2\text{EL}$~\cite{jackermeier2024dual} adapts the
relational model of BoxE~\cite{abboud2020boxe}, which also allows for
capturing role composition and role subsumption.
BoxEL~\cite{xiong2022faithful} develops an alternative way to
elucidate relations by affine transformations which solves the problem
of concept embedding size incompatibility.
In~\cite{lacerda2023strong} a proper non-convex geometric
interpretation for concepts, roles and individuals is introduced first
through mapping of interpretation domain elements into binary vectors,
and then the convex hulls of constructed regions are considered.  This
method's applicability remains an open
question since it lacks implementation and empirical evaluation.
\cite{ozcep2023embedding} is another method that is mainly described
from a theoretical perspective. It uses axis-aligned cones for concept
interpretations in $\mathcal{ALC}$ ontologies.  As opposed to
previously discussed works, this method builds partial models: as an
example, if for some individual $a$ only assertion axioms of
$(C \sqcup D)(a)$ and $C(a)$ are presented within the ontology and
neither $D(a)$ nor $(\neg D)(a)$ can be proven, the embedding
leverages multiple interpretations since there are several dissimilar
ways to interpret $a$ within this geometric framework.  There are some
other methods tailored to encode $\mathcal{ALC}$ theories, using fuzzy
sets~\cite{tang2022falcon} or ordered vector spaces \cite{zhapa2023cate}.

From the perspective of the expressivity of the DL embedded, many
geometric models focus on encoding the constructs of
$\mathcal{EL}^{++}$. ELEmbeddings~\cite{kulmanov2019embeddings},
ELBE~\cite{peng2022description} and BoxEL~\cite{xiong2022faithful}
work with the $\mathcal{EL}$ fragment enriched with nominals omitting
role inclusions.  Subsumption axioms are interpreted as the
containment of one geometric region within another one.
EmEL++~\cite{mondal2021emel++} and
$\text{Box}^2\text{EL}$~\cite{jackermeier2024dual} include objective
functions for role inclusion and role chain axioms adopting
geometrical containment for role interpretations: in
$\text{Box}^2\text{EL}$, a box inclusion loss is used towards boxes
that represent the head and tail parts of a role for role inclusion
and chain axioms; in EmEL++~\cite{mondal2021emel++}, the role
hierarchy is interpreted via establishing partial order for vectors.
\cite{lacerda2023strong} provides detailed information about each
element in the interpretation domain (including to what individual it
corresponds, in which concepts it is contained and how it is related to
other elements); while this approach can be applied to
$\mathcal{ALC}$, it requires to store sparse vectors of size $|\bm{N}_{I}| + |\bm{N}_{C}| + |\bm{N}_{R}|\cdot |\Delta^{\mathcal{I}}|$, where $\Delta^{\mathcal{I}}$ is the interpretation domain.
Other methods
under consideration~\cite{ozcep2023embedding}\cite{tang2022falcon}\cite{zhapa2023cate} target the full expressivity of $\mathcal{ALC}$; in
particular, the negation of a concept is incorporated as a membership function of $(\neg C)^{\mathcal{I}}$ assigned to all individuals via fuzzy negation operator applied to the corresponding membership function of $C^{\mathcal{I}}$~\cite{tang2022falcon},
by concept lattice enrichment~\cite{zhapa2023cate}, or using polar operators
\cite{ozcep2023embedding}.

An ontology may have an ABox and an RBox, besides a TBox.  Some
embedding methods do not distinguish between individuals and concepts.
They eliminate ABox and work solely with TBox
axioms~\cite{kulmanov2019embeddings}\cite{peng2022description}\cite{zhapa2023cate}
by translating assertional axioms $C(a)$ and $r(a, b)$ into TBox
axioms $\{a\} \sqsubseteq C$ and $\{a\} \sqsubseteq \exists r.\{b\}$,
respectively. Those methods allowing for role chain and role
inclusion~\cite{mondal2021emel++}\cite{jackermeier2024dual} incorporate
additionally RBox constructs.  In the case that the ABox is retained,
assertional axioms are either embedded alongside with terminological
axioms~\cite{tang2022falcon}\cite{xiong2022faithful}\cite{jackermeier2024dual} or placed into the
latent space after the model of TBox is
constructed~\cite{ozcep2023embedding}. In contrast with other
approaches, the method in \cite{lacerda2023strong} does not explicitly
represent the geometric interpretations of individuals, roles and
concepts for $\mathcal{ELH}$ ontologies: it constructs a binary vector $\mu(d)$ of size $|\bm{N}_{I}| + |\bm{N}_{C}| + |\bm{N}_{R}|\cdot |\Delta^{\mathcal{I}}|$
associated with each domain element $d$ such that $\mu(d)[a] = 1$ if $d = a^{\mathcal{I}}$ (for individuals), $\mu(d)[A] = 1$ if $d \in A^{\mathcal{I}}$ (for classes) and $\mu(d)[r, e] = 1$ if $(d, e) \in r^{\mathcal{I}}$ (for relations) ($\mu(d)[a] = 0$, $\mu(d)[A] = 0$, $\mu(d)[r, e] = 0$ otherwise). 

In order to show that learned embeddings construct a logical geometric
model of a given ontology, many methods provide an explicit proof. In
most cases, theoretical results state that if the optimization
objective converges to a certain value during training and some other
conditions are satisfied, the theory has a
model~\cite{jackermeier2024dual}\cite{kulmanov2019embeddings}\cite{tang2022falcon}\cite{xiong2022faithful},
and this model is constructed through the optimization process; in
this sense, faithfulness (machine learning) holds. Sometimes, authors
refer to this property as {\it
  soundness}~\cite{jackermeier2024dual}\cite{kulmanov2019embeddings}\cite{tang2022falcon}. For $\mathcal{ELH}$ ontology embeddings~\cite{lacerda2023strong} and axis-aligned cone embeddings~\cite{ozcep2023embedding} our definition of faithfulness is not applicable since in these works only interpretation domains and interpretation functions are introduced without describing optimization process which approximates geometric models. Definitions of strong and weak faithfulness discussed in these works are properties of an interpretation function $\mathcal{I}$.

As for works that do not provide theoretical proofs,
EmEL++~\cite{mondal2021emel++} cannot faithfully capture the role
hierarchy although the corresponding loss is present: the $r \sqsubseteq s$ loss is symmetric with respect
to $r$ and $s$, so $r \sqsubseteq s$ necessarily implies
$s \sqsubseteq r$; the same holds for role chains
$r_1 \circ r_2 \sqsubseteq s$ and
$r_2 \circ r_1 \sqsubseteq s$.
ELBE~\cite{peng2022description} has faithfulness property which relies on justifications discussed in~\cite{kulmanov2019embeddings} since each of its normal form losses is a multi-dimensional version of corresponding ELEmbeddings
objectives; faithfulness (machine learning) in this particular case
can be formulated as follows (using notations from the paper): if
$margin$ is a vector with non-positive components and the total loss
is equal to $0$ then the model is
constructed. CatE~\cite{zhapa2023cate} does not generate models, yet
it can be considered as faithful with respect to the lattice of ontology concepts
in the sense that the
transformation from the ontology to the lattice is total and
injective, and the embedding preserves lattice structure.

Based on the analysis of these geometric models for complex ontology embedding, we have the following discussions:
\begin{itemize}[leftmargin=*]
\item These models adopt highly interpretable balls and axis-aligned boxes as well as fuzzy sets and order embeddings,  
supporting constructs that are not covered by simple ontology embedding models~\cite{mondal2021emel++}\cite{jackermeier2024dual}\cite{ozcep2023embedding}\cite{tang2022falcon}\cite{lacerda2023strong}, and contributing more fine-grained ontology embeddings~\cite{xiong2022faithful}\cite{zhapa2023cate}\cite{peng2022description}.
\item In terms of expressivity, these models are tailored to main DL constructs of either  $\mathcal{EL}^{++}$~\cite{kulmanov2019embeddings}\cite{mondal2021emel++}\cite{peng2022description}\cite{jackermeier2024dual}\cite{xiong2022faithful}\cite{lacerda2023strong} or $\mathcal{ALC}$~\cite{ozcep2023embedding}\cite{tang2022falcon}\cite{zhapa2023cate}. As one potential future direction, we can mention targeting the full expressivity of $\mathcal{EL}^{++}$, motivated by real-world applications (e.g., many ontologies in biomedical domain) and stronger reasoning capabilities.
\item Applicability to real-world scenarios remains an open question for some embedding methods~\cite{ozcep2023embedding}\cite{lacerda2023strong}, requiring more evaluation on real-world ontologies and tasks.
\end{itemize}

\subsection{Embedding Ontology with Literals}\label{sec:oel}
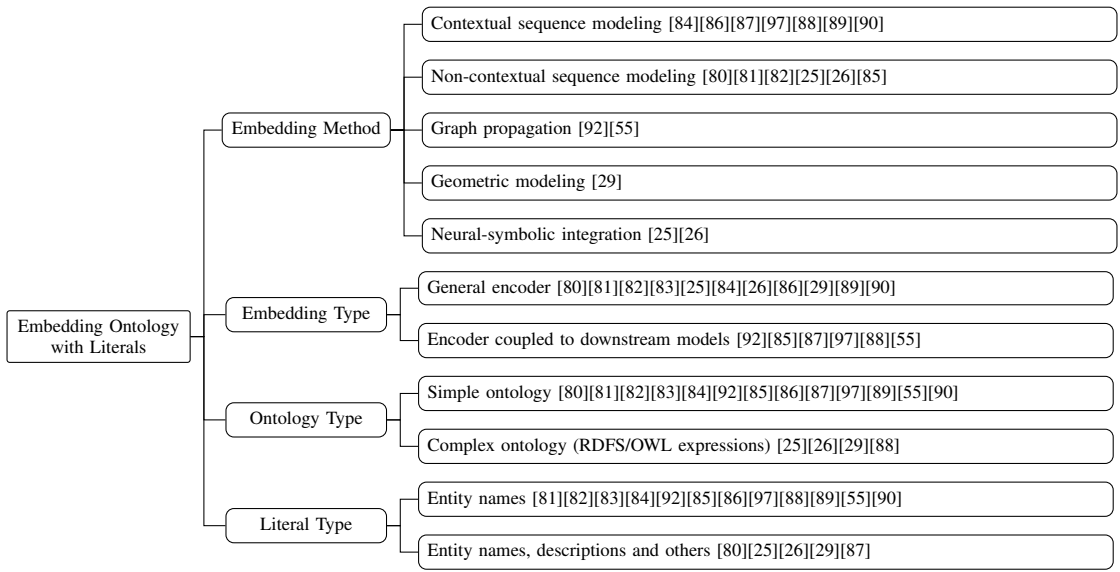
\begin{figure*}[!ht]
     \centering
\begin{forest}
    for tree={forked edges, grow'=0, draw, rounded corners, node options={align=center,}, text width=2.7cm, font=\scriptsize},
    [Embedding Ontology with Literals, 
    for tree={fill=white!45, parent, font=\scriptsize}
        [Embedding Method, for tree={fill=white!45, child, text width=2cm, node options={align=center,}, font=\scriptsize}
            [Contextual sequence modeling \cite{liu2020concept}\cite{liu2021self}\cite{zhang2022ontoprotein}\cite{he2022bertmap}\cite{chen2023contextual}\cite{gosselin2023sorbet}\cite{he2024language}, fill=white!30, grandchild, text width=9cm, font=\scriptsize]
            [Non-contextual sequence modeling \cite{xiang2015ersom}\cite{jayawardana2017deriving}\cite{kolyvakis2018deepalignment}\cite{smaili2019opa2vec}\cite{chen2021owl2vec}\cite{nguyen2021biomedical}, fill=white!30, grandchild, text width=9cm, font=\scriptsize]
            [Graph propagation \cite{hao2021medto}\cite{hao2023ontology}, fill=white!30, grandchild, text width=9cm, font=\scriptsize]
            [Geometric modeling \cite{geng2021ontozsl}, fill=white!30, grandchild, text width=9cm, font=\scriptsize]
            [Neural-symbolic integration \cite{smaili2019opa2vec}\cite{chen2021owl2vec}, fill=white!30, grandchild, text width=9cm, font=\scriptsize]
        ]
        [Embedding Type, for tree={fill=white!45,child, node options={align=center,}, font=\scriptsize}
            [General encoder \cite{xiang2015ersom}\cite{jayawardana2017deriving}\cite{kolyvakis2018deepalignment}\cite{dong2019imposing}\cite{smaili2019opa2vec}\cite{liu2020concept}\cite{chen2021owl2vec}\cite{liu2021self}\cite{geng2021ontozsl}\cite{gosselin2023sorbet}\cite{he2024language}, fill=white!30, grandchild, text width=9cm, font=\scriptsize]
            [Encoder coupled to downstream models \cite{hao2021medto}\cite{nguyen2021biomedical}\cite{zhang2022ontoprotein}\cite{he2022bertmap}\cite{chen2023contextual}\cite{hao2023ontology}, fill=white!30, grandchild, text width=9cm, font=\scriptsize]
        ]
        [Ontology Type, for tree={fill=white!45, child, node options={align=center,}, font=\scriptsize}
            [Simple ontology \cite{xiang2015ersom}\cite{jayawardana2017deriving}\cite{kolyvakis2018deepalignment}\cite{dong2019imposing}\cite{liu2020concept}\cite{hao2021medto}\cite{nguyen2021biomedical}\cite{liu2021self}\cite{zhang2022ontoprotein}\cite{he2022bertmap}\cite{gosselin2023sorbet}\cite{hao2023ontology}\cite{he2024language}, fill=white!30, grandchild, text width=9cm, font=\scriptsize]
            [Complex ontology (RDFS/OWL expressions) \cite{smaili2019opa2vec}\cite{chen2021owl2vec}\cite{geng2021ontozsl}\cite{chen2023contextual}, fill=white!30, grandchild, text width=9cm, font=\scriptsize]
        ]
        [Literal Type, for tree={fill=white!45, child, node options={align=center,}, font=\scriptsize}
            [Entity names \cite{jayawardana2017deriving}\cite{kolyvakis2018deepalignment}\cite{dong2019imposing}\cite{liu2020concept}\cite{hao2021medto}\cite{nguyen2021biomedical}\cite{liu2021self}\cite{he2022bertmap}\cite{chen2023contextual}\cite{gosselin2023sorbet}\cite{hao2023ontology}\cite{he2024language}, fill=white!30, grandchild, text width=9cm, font=\scriptsize]
            [Entity names{,} descriptions and others \cite{xiang2015ersom}\cite{smaili2019opa2vec}\cite{chen2021owl2vec}\cite{geng2021ontozsl}\cite{zhang2022ontoprotein}, fill=white!30, grandchild,text width=9cm, font=\scriptsize ]
        ]
    ]
\end{forest}
\caption{Dimensions, their values and corresponding works of embedding ontology with literals.}
\label{fig:ontolitdim}
\end{figure*}

We analyse those works for embedding ontologies with literals from
four dimensions --- embedding method, embedding type, ontology type
and literal type.  Their potential values as well as the corresponding
works are shown in Figure~\ref{fig:ontolitdim}.  For embedding type, we divide the methods into two kinds: general encoders which are
separately trained and applicable to different downstream tasks, and
coupled encoders which are jointly trained with some other attached
models often for feature learning and cannot be applied without these
models.  For literal type, we classify the works into those only
use entity name information, and
those use entity name information and other literals such as long descriptions.

Most methods for embedding ontologies with literals adopt the solution
of sequential modeling demonstrated in Figure~\ref{fig:gm}.  The
biggest challenge of this solution lies in its first step ---
extracting literal-injected sequences from the ontology.
Some methods such as ERSOM \cite{xiang2015ersom}, DeepAlignment
\cite{kolyvakis2018deepalignment}, $\mathcal{N}$-ball Embeddings
\cite{dong2019imposing}, SapBERT \cite{liu2021self} and HiT
\cite{he2024language} directly use the literals of an entity like names and descriptions as its sequences to learn its
embedding.  Such \textit{literal-alone sequences} miss the formal
semantics of the entity, and therefore, some other methods such as
BERTSubs \cite{chen2023contextual}, SORBERT \cite{gosselin2023sorbet}
and \cite{liu2020concept} extract \textit{context-augmented literal
  sequences} by first exploring the serialization of an entity's
contexts such as the axioms this entity is involved and its
neighbourhood in a graph transformed from the ontology, and then
textualisating the entity sequences by replacing (a part) of the
entities by their names.  For deeper integration of the literals and
the formal semantics, some more complex strategies, such as merging
corpora of different kinds of sequences, and concatenating a literal-alone sequence and a context-augmented literal sequence of an
entity for a \textit{hybrid literal sequence}, have also been proposed in OPA2Vec \cite{smaili2019opa2vec} and
$\text{OWL2Vec}^{*}$ \cite{chen2021owl2vec}.

In the second step of this sequence learning solution, many methods directly train an encoder from the extracted sequences by
unsupervised  learning.  ERSOM \cite{xiang2015ersom}
trains a stacked auto-encoder which is a neural network with
several hidden layers; DeepAlignment
\cite{kolyvakis2018deepalignment}, OPA2Vec \cite{smaili2019opa2vec}
and $\text{OWL2Vec}^{*}$ \cite{chen2021owl2vec} directly train a Word2Vec model;
\cite{liu2020concept} trains a biomedical variant of BERT. 
There are also some methods trying to utilize some external tasks and data for training. 
SORBERT \cite{gosselin2023sorbet} trains a sentence transformer with a siamese network architecture by minimizing the
distance between matched concepts from two ontologies; SapBERT
\cite{liu2021self} trains a biomedical BERT by minimizing the distance
between a mention from text and its matched entity in an ontology; HiT
\cite{he2024language} re-trains variants of BERT by retaining the
subsumption relationships of an ontology in a Poincar\'e ball.

The encoders by the above two kinds of methods are usually quite
general, i.e., they are applicable to different tasks and/or different
other ontologies beyond the ones for training, and their embeddings can be fed to
different other models.  On the other hand, some encoders for ontology
embedding are trained in conjunction with some downstream models by
specific tasks.  BERTSubs \cite{chen2023contextual} fine-tunes a BERT
model and an attached classifier with concept subsumptions; OntoProtein \cite{zhang2022ontoprotein}
trains a BERT model in conjunction with learning the embeddings of
proteins and the Gene Ontology.  Such learned ontology embeddings
usually have limited generality, and can only be applied with another jointly trained model.

Besides sequence learning, OntoZSL \cite{geng2021ontozsl} embeds an
RDFS ontology with entity names and descriptions by a geometric
modeling method which extends TransE with additional losses between
entity representations and literal representations.  MEDTO
\cite{hao2021medto} and \cite{hao2023ontology} both use Graph Neural
Networks to learn concept features via propagation over concept
hierarchies for minimizing distances between matched
concepts.  OPA2Vec \cite{smaili2019opa2vec} and $\text{OWL2Vec}^{*}$
\cite{chen2021owl2vec} also adopt neural-symbolic
integration by employing OWL reasoners for inferring hidden axioms for
augmenting the sequence extraction.

With the above analysis on the current works, we have the following
discussion for ontology embedding with literals:
\begin{itemize}[leftmargin=*]
\item Sequence modeling is the most widely used general solution. Among these works, using
  contextual word embedding often leads to better performance in downstream
  tasks, but it requires more future efforts
  to train general encoders.

\item Geometric modeling has been explored only in \cite{geng2021ontozsl}. However, it can train
  embeddings with higher interpretability, through which we can
  understand how knowledge are inferred and learn the impact of semantics of literals.

\item Current works focus on textual literals such as short phrases of names and long text of
  descriptions. 
  Literals of other data types such as numbers are simply regarded as plain text. More future efforts are worthy while to deal
  with literals of different types especially
  images for more comprehensive multi-modal ontology embedding.

\end{itemize}

\subsection{Embedding Ontology with KG}\label{sec:oekg}

\begin{figure*}[!ht]
     \centering
\begin{forest}
    for tree={forked edges, grow'=0, draw, rounded corners, node options={align=center}, text width=2.7cm},
    [Embedding Ontology with KG, fill=white!45, parent, text width=3.7cm, font=\scriptsize
        [Embedding Method, for tree={fill=white!45, child, text width=2cm, node options={align=center}, font=\scriptsize}
            [Point-based concept modeling \cite{hao2019universal}\cite{xiang2021ontoea}, fill=white!30, grandchild, font=\scriptsize]
            [Sphere-based concept modeling \cite{lv2018differentiating}\cite{diaz2018embeds}, fill=white!30, grandchild, font=\scriptsize]
            [Ellipsoid and box-based concept modeling \cite{yu2023geometry}\cite{huang2023concept2box}\cite{wang2024embedding}, fill=white!30, grandchild, font=\scriptsize]
        ]
        [Ontology Type, for tree={fill=white!45, child, node options={align=center}, font=\scriptsize}
            [Simple ontology \& KG \cite{lv2018differentiating}\cite{xiang2021ontoea}\cite{yu2023geometry}\cite{wang2024embedding}, fill=white!30, grandchild, font=\scriptsize]
            [Complex ontology (RDFS expressions) \& KG \cite{diaz2018embeds}\cite{hao2019universal}\cite{huang2023concept2box}, fill=white!30, grandchild, font=\scriptsize]
        ]
    ]
\end{forest}
\caption{Dimensions, their values and corresponding works of embedding ontology with KG.}
\label{fig:ontokgdim}
\end{figure*}
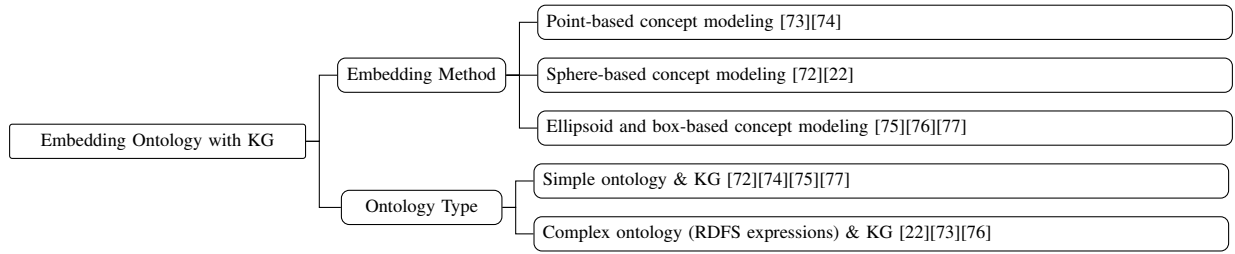

We analyze those works for embedding ontologies with KGs from two
dimensions --- embedding method and ontology type.  Note we consider
those works that use the KG for supporting ontology embedding or
jointly embed the KG and the ontology, but ignore those works that
only use ontologies for
supporting KG embedding (e.g., \cite{zhang2019iteratively}\cite{xie2016representation}). See \cite{zhang2022knowledge} for a survey of the latter.  As all the works under consideration
use geometric modeling, we make a more
fine-grained categorization according to how concepts are
modeled.  Their potential values and corresponding works are shown in Figure \ref{fig:ontokgdim}.

Some works consider joint embedding of  hierarchical concepts and a KG with relational facts \cite{lv2018differentiating}\cite{xiang2021ontoea}\cite{yu2023geometry}\cite{wang2024embedding}.  
TransC \cite{lv2018differentiating} represents each concept as a sphere and each instance as a point in the Euclidean space.  The
concept membership is then modeled by point  inclusion in sphere, and the concept subsumption is modeled by sphere inclusion. Their losses
are jointly minimized together with the translation loss of TransE that models the facts.  
For higher expressivity, some more complex geometric objects are used.  
TransEllipsoid \cite{yu2023geometry} and EIKE \cite{wang2024embedding} both model each concept as a high dimensional ellipsoid, and TransCuboid \cite{yu2023geometry} models each concept as a high dimensional box. The ellipsoid and box are both modeled by one vector for the center and another vector for the boundary (i.e., offset). TransEllipsoid, TransCuboid and EIKE use similar losses as TransC for training.
OntoEA \cite{xiang2021ontoea} jointly embeds two KGs and their ontologies.
It uses a point to represent each concept, and models not only concept subsumption, instance membership and relational facts, but also instance matching across KGs and concept disjointness.

The other works consider embedding of RDFS ontologies with KGs \cite{diaz2018embeds}\cite{hao2019universal}\cite{huang2023concept2box}.  
JOIE \cite{hao2019universal} transforms the RDFS ontology into an ontology view graph, with each concept modeled as a node, and each relation's domain concepts and range concepts connected by this relation.  
In training, triples of the ontology view graph are equally modeled as normal KG triples by a translation loss, and the instance membership is
modeled by a mapping from the instance to the concept.
Concept2Box \cite{huang2023concept2box} extends JOIE by modeling each concept as a box and by replacing the translation loss with a binary cross entropy loss.  
EmbedS \cite{diaz2018embeds} models each concept as a sphere and each relation by two spheres --- one for its domain and the other for its range, and uses distance-based losses. However, EmbedS has not been evaluated.

With the above analysis, we have the following observations
and perspectives on embedding ontology with KG:
\begin{itemize}[leftmargin=*]
\item Although the technical solutions of sequence
  modeling and graph propagation have not been exploited in the current works, through transforming the ontology and the
  KG into one graph or directly into sequences, they can be applied.
\item The current works adopt typical KG completion benchmarks for evaluation. More real-world benchmarks and complex tasks are required to full evaluation. 
  Meanwhile, some complex ontology embedding methods such as $\text{Box}^2\text{EL}$ \cite{jackermeier2024dual} are also applicable, but have not been evaluated.
\end{itemize}

\subsection{Complexity of Ontology Embedding Methods}\label{sec:complexity}

We analyze the space and time complexity of ontology embedding methods
in Table~\ref{tab:complexity}. The notation
$\bm{N}_C, \bm{N}_R, \bm{N}_I$ represents the set of concepts, role
and individual names within an ontology, respectively, and $d$ is the embedding dimension. 
Space complexity describes the number of memory units
to store embedding vectors. Most of the methods increase linearly with
$d$ because ontology entities are represented as vectors in
$\mathbb{R}^d$. However, some methods, like the one in
\cite{ozcep2023embedding} exhibit a quadratic space complexity,
\bigo{d^2}, because role entities are stored as matrices in
$\mathbb{R}^{d\times d}$. Particular cases include
\cite{lacerda2023strong}, where
$d= \numc + \numi + \numr \cdot \deltai$ and each element in the
domain $\Delta^\mathcal{I}$ is mapped to a binary vector $v$ of length
$d$ and \cite{zhapa2023cate}, where complexity scales linearly with
the number of operators $op$ in ontology axioms. Operators are symbols
$\sqcap$, $\sqcup$, $\neg$, $\exists$, $\forall$ found in ontology
axioms. Since $\mathcal{ALC}$ allows arbitrarily long axioms, we
assume that
$\bigo{d\cdot op} \gg \bigo{d \cdot (\numc + \numr + \numi)}$ in most
cases.

Time complexity $\bigo{d}$ involves linear operations such as scalar
multiplication or element-wise vector operations. In contrast, methods
with complexity $\bigo{d^2}$ involve quadratic operations which
usually take the form of matrix-vector multiplications. A particular
case is \cite{tang2022falcon}, where the complexity is
$\bigo{d^2 \cdot \deltai}$ because concept descriptions involving
existential or universal restrictions involve aggregation operations
over all elements in the domain $\Delta^{\mathcal{I}}$.

Ontology embeddings with literals are particularly different as the set of entities include not only concepts, individuals and roles, but also a potentially large
vocabulary from descriptions, labels or external
documents. Therefore, the complexity analysis incorporates the
vocabulary size $|V|$ and we assume that
$|V| \geq \numc + \numi + \numr$. Methods in \cite{smaili2019opa2vec} and
\cite{chen2021owl2vec} incorporate Word2Vec, whose space and time
complexity are $\bigo{d \cdot |V|}$ and
$\bigo{c \cdot d \cdot log_2(|V|)}$, respectively, where $c$ represents
the context size in Word2Vec. The complexity of methods that implement
random walks~\cite{chen2021owl2vec}\cite{hao2023ontology} include
parameters such as number of walks $w$ and walk length $l$. We use the
parameter $L$ to represent the number of layers in a neural networks
used in methods such as \cite{xiang2015ersom} and
\cite{hao2021medto}. Finally, we do not include
methods that involve tranining/fine-tuning language models because their large
number of parameters and training time can obscure the complexity
analysis.

\begin{table*}[t]
  \centering
  \begin{tabular*}{\linewidth}{@{\extracolsep\fill}lll@{\extracolsep\fill}}
    \toprule
    Method                                                       & Space Complexity                                               & Time Complexity \\
    \midrule
                                                                 & \multicolumn{2}{c}{Simple Ontology} \\ \cmidrule{2-3}
    Order Embeddings~\cite{vendrov2015order}                     & \bigo{d \cdot \numc}                                           & \bigo{d}\\
    Poincar\'e Embeddings~\cite{nickel2017poincare}              & \bigo{d \cdot \numc}                                           & \bigo{d} \\
    Hyperbolic Entailment Cones~\cite{ganea2018hyperbolic}       & \bigo{d \cdot \numc}                                           & \bigo{d} \\
    Box Lattices~\cite{vilnis2018probabilistic}                  & \bigo{2d \cdot \numc}                                          & \bigo{d}\\
    Density Order Embeddings~\cite{athiwaratkun2018hierarchical} & \bigo{2d \cdot \numc }                                         & \bigo{d} \\
    Smooth Boxes~\cite{li2018smoothing}                          & \bigo{2d \cdot \numc}                                          & \bigo{d} \\
    Joint Hierarchies with Boxes~\cite{patel2020representing}    & \bigo{4d \cdot \numc}                                          & \bigo{d}\\
    Gumbel Box Embeddings~\cite{dasgupta2020improving}           & \bigo{2d \cdot \numc}                                          & \bigo{d} \\
    HBE~\cite{pan2021hyperbolic}                                 & \bigo{d \cdot (\numc + \numr)}                                 & \bigo{d^2} \\
    HYPON~\cite{li2021hypon}                                     & \bigo{d\cdot \numi}                                            & \bigo{d^2}\\
    HyperExpan\cite{ma2021hyperexpan}                            & \bigo{d \cdot \numc}                                           & \bigo{d^2} \\ \cmidrule{2-3}

                                                                 & \multicolumn{2}{c}{Complex Ontology} \\ \cmidrule{2-3} 
    ELEmbeddings~\cite{kulmanov2019embeddings}                   & \bigo{d\cdot (\numc + \numi + \numr) + \numc + \numi}                  & \bigo{d} \\
    EmEL++~\cite{mondal2021emel++}                               & \bigo{d\cdot (\numc + \numi + \numr) + \numc + \numi}                  & \bigo{d} \\
    ELBE~\cite{peng2022description}                              & \bigo{d\cdot (2\numc + 2\numi + \numr)}                         & \bigo{d} \\
    BoxEL~\cite{xiong2022faithful}                               & \bigo{d\cdot (2\numc +  \numi + 2\numr)}                       & \bigo{d} \\ 
    Box$^2$EL~\cite{jackermeier2024dual}                         & \bigo{d\cdot (3\numc + 2\numi + 4\numr)}                       & \bigo{d} \\
    Convex/Non-convex Regions~\cite{lacerda2023strong}           & \bigo{\deltai \cdot (\numc + \numi + \numr \cdot \deltai)}     & \bigo{\deltai ^5}\\
    Al-Cones~\cite{ozcep2023embedding}                           & \bigo{d \cdot (\numc + \numi) + d^2 \cdot \numr}               & \bigo{d^2}\\
    FALCON~\cite{tang2022falcon}                                 & \bigo{d\cdot ( {\numc + \numi + |\mathbf{N}_{I_e}| + \numr}) } & \bigo{d^2 \cdot \deltai} \\
    CatE~\cite{zhapa2023cate}                                    &  \bigo{d \cdot op}                                             & \bigo{d} \\ 
    \cmidrule{2-3}
                                                                 & \multicolumn{2}{c}{Ontology with KGs}\\ \cmidrule{2-3}
    TransC \cite{lv2018differentiating}                          & \bigo{d\cdot (\numc + \numi + \numr) + \numc}                  & \bigo{d}\\
    EmbedS~\cite{diaz2018embeds}                                 &  \bigo{d\cdot (\numc + \numi + 2\numr) + \numc + \numr}        & \bigo{d}\\ 
    JOIE~\cite{hao2019universal}                                 &  \bigo{d_1\cdot (\numc + \numr) + d_2\cdot (\numi + \numr)}    & \bigo{d_1^2 + d_1d_2}\\ 
    OntoEA \cite{xiang2021ontoea}                                & \bigo{d\cdot (\numc + \numi + \numr)}                          & \bigo{d^2}\\
    TransEllipsoid/TransCuboid~\cite{yu2023geometry}             & \bigo{d\cdot (2\numc + \numi + \numr)}                         & \bigo{d}\\
    Concept2Box~\cite{huang2023concept2box}                      & \bigo{d \cdot (2\numc + 3\numr + \numi)}                       & \bigo{d\cdot d_{BERT}}\\
    EIKE~\cite{wang2024embedding}                                & \bigo{d\cdot (2\numc + \numi) + \numr}                         & \bigo{d^2} \\
    \cmidrule{2-3}
                                                                 & \multicolumn{2}{c}{Ontology with Literals} \\ \cmidrule{2-3}
    ERSOM~\cite{xiang2015ersom}                                  & \bigo{|V| \cdot (\numc + \numr + \numi)}                       & \bigo{L \cdot d^2} \\
    DeepAlignment~\cite{kolyvakis2018deepalignment}              & \bigo{d \cdot |V|}                                             &  \bigo{d} \\ 
    Category Trees~\cite{dong2019imposing}                       & \bigo{\log_b(|V|)}                                             & \bigo{\log_b(|V|)}\\
    MEDTO~\cite{hao2021medto}                                    & \bigo{d \cdot \numc}                                           & \bigo{L\cdot d^2} \\
    Semantic/Structural Embeddings~\cite{hao2023ontology}        & \bigo{ w \cdot l \cdot |V|+  d\cdot |V|}                       & \bigo{w \cdot l \cdot |V| +  c \cdot d \cdot \log_2(|V|) + d^2}\\
    OPA2Vec~\cite{smaili2019opa2vec}                             & \bigo{d\cdot |V|}                                              & \bigo{c \cdot d \cdot \log_2(|V|)}\\
    OWL2Vec*~\cite{chen2021owl2vec}                              & \bigo{ w \cdot l \cdot |V|+  d\cdot |V|}                       & \bigo{w \cdot l \cdot |V| +  c \cdot d \cdot \log_2(|V|)}\\
    \bottomrule
  \end{tabular*}
  \caption[Complexity of Ontology Embedding Methods]{Time and Space Complexity of Ontology Embedding Methods.}
  \label{tab:complexity}
\end{table*}

\revision{
\section{Ontology Embeddings for Knowledge Engineering and Machine Learning}\label{sec:application}
}

\subsection{Knowledge Engineering}

\subsubsection{Ontology Matching}
Given two ontologies $\mathcal{O}_1$ and $\mathcal{O}_2$, ontology matching (OM) is to find out entity mappings in form of $(e_1, e_2)$, where $e_1$ and $e_2$ are from $\mathcal{O}_1$ and $\mathcal{O}_2$, respectively, with an equivalence or subsumption relationship \cite{otero2015ontology}. 
A good OM system is expected to have a high Precision for the discovered mappings and a high Recall towards the ground truth mappings.
Sometimes, some entities in one ontology are given, and for each of them, an OM system is expected to rank the entities in the other ontology such that the truly matched entity is ranked in the first position. In this situation, ranking-based metrics like MRR (Mean Reciprocal Rank) and Hits@K (K=1, 5, 10, ...) are often used for evaluation.

Traditional systems mostly use some of the following three techniques: lexical matching, graph structure matching and logical reasoning. However, they are limited in several aspects such text understanding and fusion of different semantics.
Ontology embeddings provide a promising solution to address these limitations, and thus have been applied for OM by several recent studies.
Meanwhile, OM benchmarks that are specifically developed for evaluating machine learning-based OM systems such as Bio-ML \cite{he2022machine} provide good contexts for evaluating ontology embeddings. 

Most embedding-based OM methods consider literals due to their important information. 
The early methods ERSOM \cite{xiang2015ersom} and DeepAlignment \cite{kolyvakis2018deepalignment} directly calculate the distance of two concept embeddings, while LogMap-ML \cite{chen2021augmenting} and SORBERT \cite{gosselin2023sorbet} further trains a supervised mapping classifier that uses concept embeddings as input. Both solutions exploit the embedded semantics for discovering mappings, but the improvement over the traditional systems is still limited as the embeddings are general with no specification to OM. 
For better performance, most recent OM methods including MEDTO \cite{hao2021medto}, \cite{nguyen2021biomedical}, BERTMap \cite{he2022bertmap}, BERTSubs \cite{chen2023contextual} and \cite{hao2023ontology} jointly learn task specific embeddings of an ontology with a model for matching. For example, BERTMap \cite{he2022bertmap} fine-tunes a PLM for encoding concepts with their names, using synonyms from the two ontologies and the given mappings in option, while BERTSubs \cite{chen2023contextual} fine-tunes a PLM for encoding concepts with their contexts for predicting concept subsumption mappings.

\subsubsection{Ontology Reasoning}~\label{sec:ontology_reasoning}
The embeddings of an ontology can be used to infer its missing knowledge, among which different forms of concept subsumptions such as $C \sqsubseteq D$, $C \sqcap D \sqsubseteq E$, $C \sqsubseteq \exists r. D$ and $\exists r. D \sqsubseteq C$, concept memberships, property domains and ranges are commonly considered by many current studies (e.g., \cite{jackermeier2024dual}, \cite{xiong2022faithful}, \cite{mondal2021emel++}, \cite{chen2021owl2vec}, \cite{chen2023contextual}, \cite{nickel2017poincare} and \cite{hao2019universal}).
In evaluation for concept subsumption inference, the sub-concept is often given, and a set of candidate concepts are ranked according to the score of being the super-concept, where MRR and Hits@K are often adopted for performance measurement.
Note the selection of candidate concepts can be quite flexible, depending on the benchmarking requirement. For example, they can be all the named or complex concepts that exist in the ontology, a particularly selected subset of them, or some particularly constructed complex concepts. 
The evaluation for concept membership inference is similar with an instance given and a set of candidate concepts ranked.

Meanwhile, there are two settings for inference: 
\begin{itemize}[leftmargin=*]
\item \textit{Prediction}. A small part of the axioms are splitted out from all the declared axioms of the ontology for testing, and the remaining declared axioms are used for training. The models are expected to capture more generalizable patterns for achieving better prediction performance.
\item \textit{Approximate Deductive Inference}. The declared axioms are used for training, while the entailed axioms are used for testing. This setting is often used for measuring whether the ontology embeddings have retained all the formal semantics.
\end{itemize}

Embeddings learned by different solutions are applied for ontology reasoning using different paradigms.
\textit{(i)} For embeddings by geometric modeling, axioms can usually be inferred by calculating the geometric relationship of vectors. For the embeddings that represent concepts by boxes, $C \sqsubseteq D$ can be inferred if the box of $C$ is fully inside the box of $D$. Otherwise, a score can also be calculated according to the relative volume of their overlap.
\textit{(ii)} Embeddings by sequence learning and graph propagation
can be regarded as pre-trained machine learning features and can be fed into another separately trained machine learning models like binary classifiers for concept subsumption prediction (e.g., \cite{chen2021owl2vec}\cite{liu2020concept}) and unsupervised clusters for concept clustering (e.g., \cite{ritchie2021ontology}). 

\subsubsection{Discussion}
We have the following observations for ontology embedding for knowledge engineering.
\textit{(i)} Embeddings by geometric modeling support interpretable inference, but often perform worse than embeddings by sequence modeling with literals incorporated. 
It is challenging but promising to incorporate literals in geometric modeling.
\textit{(ii)} Most evaluation assumes a part of the knowledge to infer (e.g., the sub-concept of a concept subsumption axiom) are given, but such settings still require much human support in real-life scenarios. We need more benchmarks and metrics for supporting end-to-end evaluation. 
\textit{(iii)} The application of ontology embeddings for knowledge engineering mostly lie in OM and inferring missing knowledge within ontology.
Other tasks such as entity resolution, query answering, knowledge retrieval and ontology learning from text can be explored.

\subsection{Knowledge Augmented Machine Learning}
Ontologies are able to represent information of machine learning tasks, datasets and algorithms, and thus ontology embedding can be a medium to inject domain knowledge into machine learning training or prediction.  
One typical aspect for augmentation is dealing with the sample shortage problem \cite{chen2023zero}\cite{chen2021knowledge}.  
In this part, we introduce a case study of using ontology embeddings for zero-shot learning (ZSL) which typically refers to a machine learning classification task\footnote{In machine learning classification, the output is often called class. To distinguish it with class in ontology, we call the output  classification label or label in brief.} with some or all of its testing labels unseen in training \cite{xian2017zero}\cite{kulmanov2022deepgozero}.
%
The model is expected to have high accuracy on testing samples of both seen and unseen labels.

Ontology-aware ZSL requires to construct or re-use an ontology that models the relationships of seen and unseen classification labels, where each label is often represented as a concept in the ontology.
For example, in animal image classification, such an ontology could represent animal taxonomies, visual characteristics, habitats, and so on.
With the ontology and its embeddings, there are mainly two paradigms among the current studies: 
\begin{itemize}[leftmargin=*]
\item \textit{Mapping-based}. In training, this paradigm learns a mapping function from the training samples to map the vector representation (e.g., image features) of the input to the ontology embedding of the output label. In prediction, the mapping function is applied to map the test input into an embedding, and the label (either seen or unseen) whose embedding is closest to this embedding is regarded as the output. It can also map the label's embedding to the input's vector, or map both to a common vector. One example work is \cite{chen2020ontology} which uses EL Embedding \cite{kulmanov2019embeddings} to embed an ontology of OWL EL for modeling labels of animals for zero-shot animal image classification. 

\item \textit{Generative}. This paradigm generates samples for an unseen label according to its embedding in the ontology, by learning a conditional generative model like Generative Adversarial Network (GAN). Thus the ZSL problem is transformed into a normal supervised learning problem.
A representative work is OntoZSL \cite{geng2021ontozsl} which embeds literal-aware ontologies for zero-shot image classification, and zero-shot KG link prediction with unseen relations.
\end{itemize}

Applying ontology embedding for machine learning sample shortage is a promising solution of neural-symbolic integration, but there is still a shortage of successful systems under deployment.
We think the limitation lies in the representation of more complex knowledge such as uncertain relationships (e.g., an image of horse may have a background of grass land, but not always), as well as the automatic construction of the ontology for a specific task. 
The corresponding solutions could be more flexible multi-modal ontologies with different kinds of literals and example instances, and more tools for knowledge integration and automatic ontology construction.

\revision{
\section{Ontology Embedding Domain Applications}\label{sec:application2}}
\subsection{Life Sciences}\label{sec:ls}

In life sciences, the development of successful
ontologies such as the Gene Ontology (GO) \cite{gene2019gene},
SNOMED-CT~\cite{donnelly2006snomed} and the Human Phenotype Ontology
\cite{Robinson_2008} has motivated the development of methods that
incorporate ontologies as a source of background knowledge. With the
rise of machine learning, ontology embedding has become a common approach to leverage the ontology.
In this part, we review works of two kinds of life science tasks which widely use ontology embeddings: \textit{(i)} protein-protein interaction, gene-disease association and protein/gene function prediction, and \textit{(ii)} healthcare predictive analysis with Electronic Health Records (EHR). 

Methods for these tasks are usually evaluated as \textit{(i)} classification systems with metrics such as Precision and Recall, \textit{(ii)} ranking systems with metrics like MRR and Hits@K, or \textit{(iii)} predictive systems with metrics like Accuracy@K (K=5,20,\dots). 
In particular cases, methods are evaluated on problem-specific metrics. For example, protein function prediction uses $F_{\max}$ which is obtained from Precision and Recall scores, and $S_{\min}$ which computes the level of uncertainty and misinformation of the predictions.

\begin{figure*}[!ht]
     \centering
\begin{forest}
    for tree={forked edges, grow'=0, draw, rounded corners, node options={align=center,}, text width=2.4cm},
    [Ontology embeddings in life sciences, for tree={fill=white!45, parent, text width=2cm, font=\scriptsize}
      [Embedding  Method, for tree={fill=white!45, child, text width=2cm, node options={align=center,},font=\scriptsize}
        [Word Embedding \cite{smaili2019opa2vec}, fill=white!30, grandchild, text width=10cm, font=\scriptsize]
        [Word Embedding + Graph-based attention \cite{choi2017gram}\cite{peng2021sequential}\cite{niu2022fusion}\cite{cheong2023adaptive}\cite{zhang2020hierarchical}\cite{song2019medical}\cite{ma2018kame}, fill=white!30, grandchild, text width=10cm, font=\scriptsize]
        [Recurrent Neural Networks~\cite{yao2023ontology}, fill=white!30, grandchild, text width=10cm, font=\scriptsize]
        [Random Walks + Word Embedding \cite{Shen_2019}\cite{Nunes_2023}\cite{embedpvp}\cite{agarwal2019snomed2vec}, fill=white!30,
        grandchild, text width=10cm, font=\scriptsize]
        [Knowledge Graph Embedding \cite{Vilela_2022}\cite{Wang_2023}\cite{zhang2022ontoprotein}\cite{Nunes_2023}\cite{embedpvp}, fill=white!30, grandchild, text width=10cm, font=\scriptsize]
        [Graph Neural Networks \cite{zhao2022}\cite{qiu2022isoform}, fill=white!30, grandchild, text width=10cm, font=\scriptsize]
        [Euclidean Balls/Boxes \cite{embedpvp} \cite{Kulmanov_2024}, fill=white!30, grandchild, text width=10cm, font=\scriptsize]
        [Hyperbolic spaces~\cite{lu2019learning}\cite{agarwal2019snomed2vec}, fill=white!30, grandchild, text width=10cm, font=\scriptsize]
        [Partial Orders \cite{li2024}, fill=white!30, grandchild, text width=10cm, font=\scriptsize] 
      ]                        
      [Application, for tree={fill=white!45, child, text width=2cm, node options={align=center,}, font=\scriptsize}
        [Protein-protein interaction \cite{smaili2019opa2vec}\cite{zhang2022ontoprotein}, fill=white!30, grandchild, text width=10cm, font=\scriptsize]
        [Gene-disease association \cite{smaili2019opa2vec}\cite{Shen_2019}\cite{Nunes_2023}\cite{embedpvp}\cite{Vilela_2022}, fill=white!30, grandchild, text width=10cm, font=\scriptsize]
        [Protein-function prediction \cite{qiu2022isoform}\cite{zhang2022ontoprotein}\cite{zhao2022}\cite{Kulmanov_2024}\cite{li2024}, fill=white!30, grandchild, text width=10cm, font=\scriptsize]
        [EHR predictive analysis \cite{song2019medical}\cite{lu2019learning}\cite{zhang2020hierarchical}\cite{peng2021sequential}\cite{niu2022fusion}\cite{yao2023ontology}\cite{agarwal2019snomed2vec}\cite{cheong2023adaptive}\cite{choi2017gram}\cite{ma2018kame}, fill=white!30, grandchild, text width=10cm, font=\scriptsize]
      ]
    ]
\end{forest}
\caption{Categorization of ontology embedding works for life sciences.}
\label{fig:lifesciences}
\end{figure*}
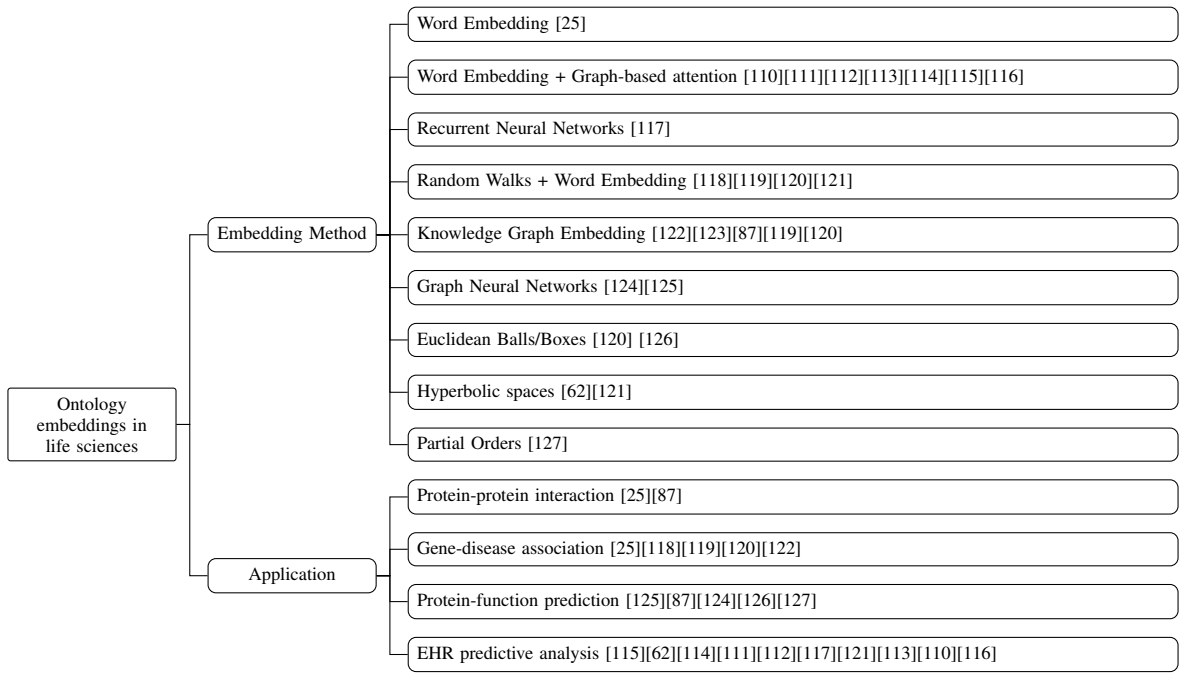

The central idea of generating ontology embeddings is to capture an entity's latent relationships with other entities. 
For example, in protein-protein interaction, whose objective is to predict if two proteins interact, methods such as \cite{smaili2019opa2vec} and \cite{zhang2022ontoprotein} link protein entities to their
corresponding functions in the Gene Ontology, and thus the generated embeddings of protein entities encode information about their functions that can be utilised for interaction prediction.
A similar idea is followed in the gene-disease association problem.
For example, in \cite{smaili2019opa2vec}, \cite{Nunes_2023},
\cite{embedpvp} and \cite{Vilela_2022}, genes and diseases 
are linked to their corresponding phenotypes in a phenotype ontology, with the goal of capturing phenotypic-related information in the embeddings. 
Other strategies, such as \cite{Shen_2019}, generate only phenotype embeddings and then compute an embedding for a gene (or disease) using their associated phenotype embeddings. 
In protein function prediction, embeddings for functions are
obtained from GO, whereas embeddings for proteins are
obtained from protein sequences, to which the use of PLMs is currently predominant \cite{esm2}\cite{prottrans}.

In healthcare, there are various EHR predictive analysis tasks such as mortality prediction, next-admission diagnosis prediction or hospital readmission prediction where hierarchical medical concepts are exploited: an ontology is compiled into a directed acyclic graph composed of its concepts and its embedding is combined with the textual information from EHRs. 
In most works including \cite{choi2017gram}, \cite{peng2021sequential}, \cite{niu2022fusion}, \cite{ma2018kame}, \cite{cheong2023adaptive}, \cite{zhang2020hierarchical}, \cite{song2019medical} and \cite{agarwal2019snomed2vec},
the graph-based attention mechanism is applied to leverage the concept hierarchies; another approach is described
in~\cite{lu2019learning} and \cite{agarwal2019snomed2vec} where concept hierarchies are embedded through hyperbolic embedding. 
Embedded medical concepts can then be used for training a Recurrent Neural Network (RNN) for sequential diagnosis prediction or mortality prediction. \cite{yao2023ontology} uses a dual RNN with co-attention and max pooling to fuse medical concept hierarchies of patient diagnoses and drugs for prescription recommendation. 
Several approaches use embeddings from multiple ontologies or multiple representations of a single ontology: \cite{cheong2023adaptive} combines multi-relational ontologies via a graph attention network for multi-relational ontology embedding; \cite{song2019medical} assigns multiple embeddings for non-leaf nodes (except for the root) of the ontology's directed acyclic graph.

The strategies that have been employed to utilise ontology embeddings for life science  can be summarised as follows:
\begin{itemize}[leftmargin=*]
\item \textit{Similarity-based strategy}: Given the trained embeddings of a pair of entities, this strategy computes the pair's score by either directly calling a similarity function or using a neural network that is additionally trained for prediction. 
  One typical embedding method that is often applied in this strategy is OPA2Vec \cite{smaili2019opa2vec}, which is to predict protein-protein interactions and gene-disease associations. 
More works that adopt this strategy include \cite{Shen_2019}, \cite{Nunes_2023} and \cite{embedpvp}. Their embeddings are also based on the sequence learning model Word2Vec, trained with random walks of the graph extracted from the ontology.

\item \textit{Graph-based strategy}: This strategy is usually based on a graph created from  one or many ontologies, applying techniques of KG embedding (KGE), neural networks, hyperbolic embedding, graph-based attention and so on. 
\cite{Vilela_2022}, \cite{Wang_2023} and \cite{zhang2022ontoprotein} frame the problem as link prediction on the graph, and address it by KGE methods such as TransE \cite{bordes2013translating} and DistMult \cite{yang2015embedding}.
OntoProtein \cite{zhang2022ontoprotein} uses the graph to enhance the training of a Protein Language Model for better protein embeddings.
PO2GO \cite{li2024} embeds a directed acyclic graph from GO using a partial-order embedding method and also adopts an additional neural network for prediction. 
Works adopting GNNs either frame the problem as node classification \cite{zhao2022} or use embeddings as partial input of a prediction module \cite{qiu2022isoform}. 
Graph-based attention models follow the original GRAM method~\cite{choi2017gram} where the final embeddings of leaf concepts are convex combinations of their own embeddings and their ancestors' embeddings. 
More works include learning interaction of hierarchical embeddings of drug and diagnoses concepts by a dual RNN co-attention model~\cite{yao2023ontology} and utilizing the Poincar\'e ball model~\cite{lu2019learning}\cite{agarwal2019snomed2vec}.

\item \textit{Model-theoretic strategy}: Since OWL ontologies are formal semantics rooted in Description Logics, this strategy aims to utilize embeddings that are generated for theories. 
In this case, the prediction problem is framed as the inference of axioms with the embeddings.  
Methods such as \cite{kulmanov2022deepgozero} for protein-function prediction and variations of \cite{embedpvp} for gene-disease associations employ a model-theoretic embedding method ELEmbeddings \cite{kulmanov2019embeddings}. 
\cite{Kulmanov_2024} further generates multiple embedding models for approximate semantic entailment.
\end{itemize}

\subsection{Other Applications}\label{sec:other}


Although ontology utilization in other domains is not as prevalent as in life science applications, partially due to the absence of well-curated or widely used domain ontologies, there are some examples of ontology exploitation with embedding, including ontology-aware classifiers for identifying research topics in scholarly articles \cite{salatino2019cso}, enhancing intelligent transportation systems \cite{ali2019fuzzy}, sentiment analysis \cite{sweidan2021sentence}, event detection \cite{deng2021ontoed} and company cointegration prediction \cite{erten2021ontology}. 
In these works, ontologies adopted are developed from scratch. 
Ontology embedding strategies in these selected studies are mostly quite simple or straightforward by applying some word or KG embedding methods, including Word2Vec in \cite{salatino2019cso} and \cite{ali2019fuzzy}, XLNet for aspect-based sentiment extraction in \cite{sweidan2021sentence}, IterE \cite{zhang2019iteratively} and Node2Vec \cite{grover2016node2vec} applied to ontology with KG in  \cite{deng2021ontoed} and \cite{erten2021ontology}. The work \cite{dassereto2020evaluating} discusses the task-dependent and task-independent evaluation of Poincar\'e disk embeddings for the GeoNames ontology \cite{wick2015geonames}.  

\section{mOWL: A Machine learning library with Ontology Embedding
  Methods}\label{sec:systems}

\begin{figure*}[!ht]
  \centering
  \includegraphics[scale=0.27]{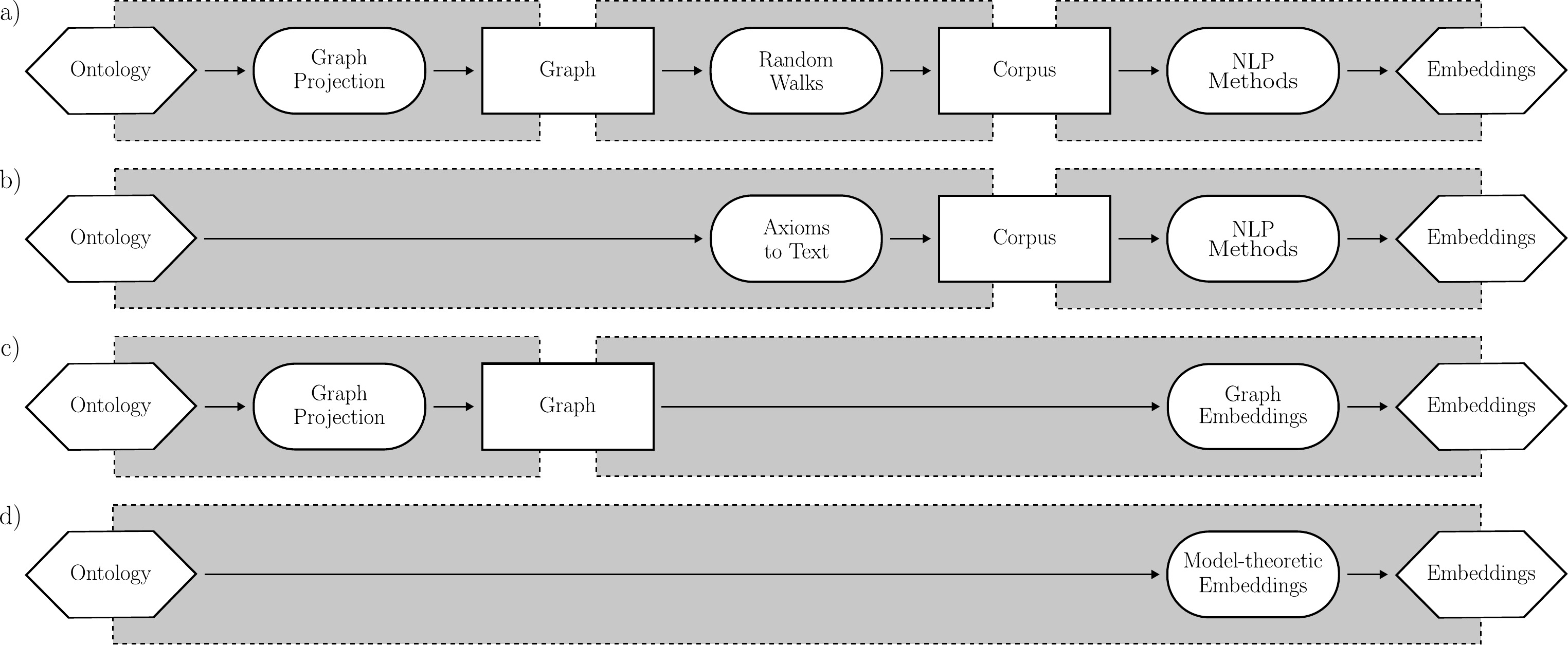}
  \caption{High-level representation of four mOWL's workflows for
    generating ontology embeddings, following the design patterns of
    neural-symbolic integration in \cite{van_Bekkum_2021}. In this
    representation, a grey box is a neural-symbolic design pattern
    that consume data, symbols or models and produce data symbols or
    models. Data and symbols are depicted with square boxes, models
    with hexahedrons and processes (such as training) with round
    boxes. Workflows
    a) and b) represent methods that transform the input ontology
    into sequences, and then learns the embeddings using NLP
    methods. These workflows are suitable for many methods of the
    sequence modeling solution.  Workflow c) represents methods that
    transform the ontology into a graph and use graph embedding
    methods. This workflow is suitable for the graph propagation
    solution.  Workflow d) directly uses an ontology's axioms to learn
    the embeddings, which are suitable methods of geometric modeling,
    especially model-theoretic methods.
  }
  \label{fig:mowl}
\end{figure*}

  Many ontology embedding works release the implementations of their methods or applications. 
  \revision{However, researchers
  often face compatibility issues between different implementations,
  spend considerable time adapting code from various sources, and
  struggle to ensure fair comparisons between methods due to differences in
  implementation.} Furthermore, there is a shortage of
easy-to-use softwares that have implemented multiple ontology
embedding methods and can support the implementation of new methods.
mOWL\footnote{\url{https://github.com/bio-ontology-research-group/mowl}}~\cite{mowl},
a library that provides functions to manipulate ontologies and use
ontologies with machine learning, aims to bridge this
gap. \revision{mOWL significantly reduces the engineering overhead for
  researchers by providing a unified API that standardizes common
  operations and allows fair comparison between methods. It enables
  researchers to focus on developing novel embedding approaches rather
  than dealing with infrastructure challenges.} mOWL provides Python
interfaces and can support the following functions: \textit{(i)}
ontology manipulation, where the OWL API~\cite{owlapi} is accessed for
ontology creation, manipulation and reasoning; \textit{(ii)} ontology
transformation, which enables extracting different graphs (such as concept hierarchies) and sequences from ontologies; \textit{(iii)}
implementation of ontology embedding methods that support ontologies of 
DL $\mathcal{EL}^{++}$ \revision{and $\mathcal{ALC}$, covering required components of many methods described in Section~\ref{sec:mc}; \textit{(iv)}
  common datasets and evaluation modules for axiom prediction and
  approximate deductive inference}.  
mOWL includes several workflows following the modular design patterns for neural-symbolic systems in \cite{van_Bekkum_2021}: 
Fig.\ref{fig:mowl} a) and b) encompass methods that transform ontology
axioms and literals into sequences and use NLP methods \revision{to
  generate embeddings}; Fig.\ref{fig:mowl} c) groups the methods that
construct graphs from ontology axioms and leverage graph propagation
methods to generate embeddings; Fig.~\ref{fig:mowl} d) groups some
model theoretic methods, especially the geometric methods targeting
DL $\mathcal{EL}^{++}$.  \revision{mOWL's modular design makes
  it particularly suitable for both research and practical
  applications. Researchers can easily conduct comparative studies
  across different embedding methods, while practitioners can
  integrate mOWL into their workflows}

\revision{To demonstrate mOWL, we use two different tasks and two benchmarks for each task. Both tasks adopt the 
  prediction setting as described in Section~\ref{sec:ontology_reasoning},
  which is to predict new axioms from the existing ontology
  background knowledge. The first task \emph{subsumption
    prediction} is to predict axioms are of the form
  $C \sqsubseteq D$ where $C,D$ are concept names. The two benchmarks used for subsumption prediction are constructed from GO and the Food Ontology\cite{chen2021owl2vec} and
  included in mOWL.  The second task  \emph{protein--protein interaction prediction} (PPI) is to determine whether two proteins interact or not
  based on their biological functions in the Gene Ontology. This task
  is formulated as prediction of axioms
  $p_i \sqsubseteq \exists \mbox{interacts\_with}. p_j$, where
  $p_i,p_j$ are instances of proteins. We tested
  PPIs for yeast and human organisms}.
\revision{Both the tasks are} framed as ranking problems, with metrics
of Mean Rank (MR), Mean Reciprocal Rank (MRR), Hits@k and AUC of ROC
curve. In Table~\ref{tab:results}, \revision{we showcase eight
  ontology embedding methods including methods that involve literals
  (OPA2Vec, OPA2Vec-NN, OWL2Vec$^*$), methods that rely on KGE (OWL2Vec$^*$-TransE), geometric methods that target DL 
  $\mathcal{EL}^{++}$ (ELEmbeddings, BoxEL, Box$^2$EL) and
  methods that target DL $\mathcal{ALC}$ (CatE).} Their implementations
in mOWL leverages a standard interface which not only eases the
implementation of ontology embedding methods, but also enables their manipulation, analysis and extension.





\begin{table*}[t]
  \centering
  \begin{tabular*}{\linewidth}{@{\extracolsep\fill}lrccccc@{\extracolsep\fill}}
    \toprule
    Method          & MR        & MRR         & H@3         & H@10        & H@100       & AUC   \\
    \midrule
    \multicolumn{7}{c}{Prediction of axioms $C \sqsubseteq D$ in GO}\\
    \midrule
    OPA2Vec         & 440       & 0.112       & 0.112       & 0.207       & 0.571       & 0.990 \\
    OPA2Vec-NN      & \fst{137} & \snd{0.167} & \snd{0.177} & \snd{0.353} & \snd{0.780}  & \fst{0.997} \\
    OWL2Vec*-Sim    & \snd{141} & \fst{0.220}  & \fst{0.245} & \fst{0.428} & \fst{0.811} & \fst{0.997} \\
    OWL2Vec*-TransE & 7446      & 0.036       & 0.037       & 0.065       & 0.161       & 0.832 \\
    ELEmbeddings    & 3279      & 0.041       & 0.045       & 0.111       & 0.330        & 0.926 \\
    BoxEL           & 6981      & 0.007       & 0.003       & 0.020       & 0.076       & 0.842 \\
    Box$^2$EL       & 3940      & 0.031       & 0.027       & 0.088       & 0.305       & 0.911 \\
    CatE            & 4548      & 0.069       & 0.081       & 0.216       & 0.433       & 0.897 \\
    \midrule
    \multicolumn{7}{c}{Prediction of axioms $C \sqsubseteq D$ in FoodOn}\\
    \midrule
    OPA2Vec         & 2094      &0.081        &0.082        &0.136        &0.349        &0.926 \\
    OPA2Vec-NN      & \fst{284} & \snd{0.112} & 0.109       & \snd{0.253} & \snd{0.645} & \fst{0.990} \\
    OWL2Vec*-Sim    & \snd{433} & \fst{0.233} & \fst{0.267} & \fst{0.442} & \fst{0.731} & \snd{0.985} \\
    OWL2Vec*-TransE & 7909      & 0.048       & 0.052       & 0.078       & 0.145       & 0.719 \\
    ELEmbeddings    & 3088      & 0.081       & 0.102       & 0.168       & 0.293       & 0.891 \\
    BoxEL           & 3790      & 0.028       & 0.035       & 0.078       & 0.192       & 0.866 \\
    Box$^2$EL       & 4655      & 0.039       & 0.051       & 0.118       & 0.221       & 0.835 \\
    CatE            & 4542      & 0.084       & \snd{0.146} & 0.197       & 0.297       & 0.839 \\
    \midrule
    \multicolumn{7}{c}{Prediction of PPI (yeast) axioms of the form $p_i \sqsubseteq  \exists \mbox{interacts\_with}. p_j$}\\
    \midrule
    OPA2Vec         & 396       & 0.061       & 0.051       & 0.128       & 0.543       & 0.935 \\
    OPA2Vec-NN      & \fst{172} & 0.144       & 0.147       & 0.326       & 0.777       & \fst{0.971} \\
    OWL2Vec*-Sim    & 197       & 0.149       & 0.154       & 0.301       & 0.730        & 0.967 \\
    OWL2Vec*-TransE & 219         & \fst{0.190} & \fst{0.203} & \snd{0.402} & \snd{0.793} & 0.964 \\
    ELEmbeddings    & 289         & 0.101       & 0.094       & 0.252       & 0.730       & 0.952 \\
    BoxEL           & 231         & 0.037       & 0.021       & 0.073       & 0.551       & 0.962 \\
    Box$^2$EL       & \snd{188}   & \snd{0.167} & \snd{0.190} & \fst{0.435} & \fst{0.805} & \snd{0.969} \\
    CatE            & 259         & 0.043       & 0.025       & 0.093       & 0.563       & 0.957 \\
    \midrule
    \multicolumn{7}{c}{Prediction of PPI (human) axioms of the form $p_i \sqsubseteq  \exists \mbox{interacts\_with}. p_j$}\\
    \midrule
    OPA2Vec         & 678       & 0.080       & 0.071       & 0.177       & 0.594       & 0.961 \\
    OPA2Vec-NN      & \fst{390} & 0.136       & 0.139       & 0.285       & \snd{0.692} & \fst{0.978} \\
    OWL2Vec*-Sim    & 568       & 0.131       & 0.140       & 0.283       & 0.639       & 0.967 \\
    OWL2Vec*-TransE & 477       & \fst{0.173} & \fst{0.187} & \fst{0.357} & \fst{0.717} & 0.972 \\
    ELEmbeddings    & 812       & 0.081       & 0.075       & 0.175       & 0.573       & 0.953 \\
    BoxEL           & \snd{411} & 0.038       & 0.021       & 0.079       & 0.564       & \snd{0.976} \\
    Box$^2$EL       & 564       & \snd{0.163} & \snd{0.175} & \snd{0.336} & 0.683       & 0.967 \\
    CatE            & 492         & 0.059       & 0.043       & 0.136       & 0.629       & 0.972 \\
    \bottomrule
  \end{tabular*}
  \vspace{-0.1cm}
  \caption[Results on axiom prediction.]{\revision{Evaluation of
      ontology embedding methods in subsumption prediction
      and protein--protein interacion prediction, with the
      dataset and models implemented in mOWL.We report filtered metrics
      that exclude axioms in the
      training set.}}
  \label{tab:results}
\end{table*}


\section{Challenges and Future Directions}\label{sec:challenge}
Although many studies have been done, ontology embedding is still a relatively new direction, and there are several challenges that prevent ontology embedding from having more real-world applications. We believe more future works are required in at least the following aspects.

\noindent\textbf{Efficient Geometric Modeling for Complex Ontologies}.
On the one hand, quite a few geometric modeling methods in the Euclidean space have been proposed, but they are mostly limited to the main features (constructs) of DL $\mathcal{EL}^{++}$ and $\mathcal{ALC}$. Many other features like the at-least and at-most restrictions have not been explored, not to mention modeling the complete formal semantics of an arbitrary OWL ontology.
Thus, we need to extend these methods to support more features that are used in real-world ontologies, with faithfulness in option.
%
On the other hand, the main contents of real-world ontologies are usually the hierarchical concepts, but their modeling in the Euclidean space (e.g., by high dimensional boxes) is much less efficient (i.e., requires many more parameters to learn) than their modeling in some hyperbolic spaces such as Poincar{\'e} ball in which the distance of points increases exponentially as they get closer to the boundary \cite{nickel2017poincare}.
There have been several studies to explore geometric modeling in hyperbolic spaces for ontology embedding, such as \cite{nickel2017poincare}, \cite{ma2021hyperexpan}, \cite{ganea2018hyperbolic} and \cite{pan2021hyperbolic}, but they mostly only embed concept hierarchies.
Geometric modeling in hyperbolic spaces for some other ontology features deserves higher attention.

\noindent\textbf{Utilising and Supporting Large Language Models (LLMs)}.
LLMs like the GPT series and the Llama series  have shown great success in understanding not only natural language but also images and (semi-)structured data  \cite{chang2024survey}\cite{touvron2023llama}. 
A promising idea is to embed ontologies, especially those with literals, using LLMs.
There have been some ontology embedding methods that use encoder-based language models, such as $\text{OWL2Vec}^{*}$ \cite{chen2021owl2vec} and OPA2Vec \cite{smaili2019opa2vec} which use Word2Vec, and HiT \cite{he2024language} which uses BERT-like Transformer-based encoders, but the generative LLMs are quite different as they adopt some decoder or encoder-decoder architectures.
Therefore, novel solutions, such as further pre-training and/or instruction tuning, need to be explored to apply them for ontology embedding.
Meanwhile, LLMs also suffer from several problems including hallucination, black-box and shortage of domain knowledge. 
Integrating KGs as well as other (semi-)structured data is widely regarded as a promising solution \cite{pan2024unifying}\cite{pan2023large}.
How can we use ontology embedding to integrate ontologies with LLMs? 
How to support Retrieval Augmented Generation (RAG) \cite{lewis2020retrieval} with ontology embedding for incorporating domain knowledge and reasoning?
Both questions are worthwhile for future ontology embedding exploration.

\noindent\textbf{Application in Neural-symbolic Integration and Domains}.
Currently, ontology embedding is mostly applied to construct and curate ontologies themselves, and link prediction for domains like life sciences
As a popular semantic technique, ontology has a high potential for building neural-symbolic integration \cite{hitzler2020neural}, utilizing ontology embedding for incorporating domain knowledge and reasoning capabilities in machine learning. 
Although some works have been proposed for ontology-based zero-shot and few-shot learning \cite{chen2023zero}, we believe the research in this direction is far from enough, with many topics like using ontologies for supporting meta learning (e.g., model selection) and augmenting explanation, not fully explored.
Meanwhile, more domain applications, both in and out of life sciences, should be considered for exploring the potential of ontology embeddings.
The current life science applications, which only consider link prediction of several simple relations and use formal semantics without literals, are quite simple. 
Link prediction with more complex target relations and data of different modalities can be explored using literal-aware ontology embedding methods.
Other tasks, such as generation for drug discovery and protein design \cite{bian2021generative} and natural language inference for clinical trial \cite{jullien2023semeval}, can also be explored with ontology embedding.

\noindent\textbf{Benchmarking}. This direction still lacks systematic benchmarking resources.
Ontology construction and inference are straightforward and effective for evaluating different aspects of ontology embeddings, including how much formal and informal semantics the embeddings retain.
But most of the current works limit the evaluation to concept subsumption inference and concept alignment. More complex tasks in ontology learning, either utilizing or not utilizing external resources, such as learning complex concept axioms, and inserting new concepts, have  been rarely considered.
Meanwhile, these aforementioned tasks for neural-symbolic integration and domain applications can be considered for benchmarking.

\section{Conclusion}\label{sec:conclusion}
This is a comprehensive survey of ontology embedding which is to represent knowledge of ontologies in vector spaces with their semantics (partially) retained. It gives formal definitions and properties of ontology embedding, and summarizes three different technical solutions, i.e., geometric modeling, sequence modeling and graph propagation, and categorizes the studies according to not only the methods used but also the ontologies they aim at (including simple ontology, complex ontology in OWL or RDFS, ontology with literal and ontology with KG).
Following the method part, the survey also gives a relatively complete analysis for the application of ontology embedding in knowledge engineering, life sciences and machine learning augmentation, and demonstrates a library mOWL developed by the co-authors that has implemented several typical ontology embedding methods and benchmarking resources. 
In the end, the survey discusses some potential future directions, including the interesting topics of integrating ontology embedding with LLMs.

\section*{Acknowledgement}
All the authors participated the discussion, paper writing and proof reading.
  Olga contributed the review
of complex ontology embedding (\ref{sec:coe}), 
applications in life sciences (\ref{sec:ls}) and other domains (\ref{sec:other}). Fernando contributed the complexity analysis 
of embedding methods (\ref{sec:complexity}), the review of applications
in life sciences (\ref{sec:ls}) and other domains (\ref{sec:other}), and the demonstration of mOWL
(\ref{sec:systems}).  Robert
co-led the work and contributed the definitions and properties (\ref{sec:dpoe}).  Yuan contributed the review
of simple ontology embedding (\ref{sec:soe}).
Jiaoyan co-led the work and contributed to the other parts.

This work has been funded by the EPSRC projects OntoEm (EP/Y017706/1), ConCur (EP/V050869/1) and UK FIRES (EP/S019111/1), the fundings from King Abdullah University of Science and Technology (KAUST) Office of Sponsored Research (OSR) under Award No. URF/1/4675-01-01, URF/1/4697-01-01, URF/1/5041-01-01, REI/1/5659-01-01, REI/1/5235-01-01, and FCC/1/1976-46-01, and the SDAIA-KAUST Center of Excellence in Data Science and Artificial Intelligence (SDAIA-KAUST AI).

\bibliographystyle{IEEEtran}
\bibliography{ontoem_reference}

\begin{thebibliography}{100}
\providecommand{\url}[1]{#1}
\csname url@samestyle\endcsname
\providecommand{\newblock}{\relax}
\providecommand{\bibinfo}[2]{#2}
\providecommand{\BIBentrySTDinterwordspacing}{\spaceskip=0pt\relax}
\providecommand{\BIBentryALTinterwordstretchfactor}{4}
\providecommand{\BIBentryALTinterwordspacing}{\spaceskip=\fontdimen2\font plus
\BIBentryALTinterwordstretchfactor\fontdimen3\font minus \fontdimen4\font\relax}
\providecommand{\BIBforeignlanguage}[2]{{%
\expandafter\ifx\csname l@#1\endcsname\relax
\typeout{** WARNING: IEEEtran.bst: No hyphenation pattern has been}%
\typeout{** loaded for the language `#1'. Using the pattern for}%
\typeout{** the default language instead.}%
\else
\language=\csname l@#1\endcsname
\fi
#2}}
\providecommand{\BIBdecl}{\relax}
\BIBdecl

\bibitem{guarino2009ontology}
N.~Guarino, D.~Oberle, and S.~Staab, ``What is an ontology?'' in \emph{Handbook on ontologies}, 2009, pp. 1--17.

\bibitem{raimond2010use}
Y.~Raimond, T.~Scott, S.~Oliver, P.~Sinclair, and M.~Smethurst, ``Use of semantic web technologies on the {BBC} web sites,'' in \emph{Linking Enterprise Data}.\hskip 1em plus 0.5em minus 0.4em\relax Springer, 2010, pp. 263--283.

\bibitem{harrison2021icd}
J.~E. Harrison, S.~Weber, R.~Jakob, and C.~G. Chute, ``{ICD-11}: an international classification of diseases for the twenty-first century,'' \emph{BMC Medical Informatics and Decision Making}, vol.~21, pp. 1--10, 2021.

\bibitem{dong2018challenges}
X.~L. Dong, ``Challenges and innovations in building a product knowledge graph,'' in \emph{ACM SIGKDD}, 2018, pp. 2869--2869.

\bibitem{horrocks2008ontologies}
I.~Horrocks, ``Ontologies and the semantic web,'' \emph{Communications of the ACM}, vol.~51, no.~12, pp. 58--67, 2008.

\bibitem{hendler2001semantic}
J.~Hendler, O.~Lassila, and T.~Berners-Lee, ``The semantic web,'' \emph{Scientific American}, vol. 284, no.~5, pp. 34--43, 2001.

\bibitem{mcbride2004resource}
B.~McBride, ``The resource description framework {(RDF)} and its vocabulary description language {RDFS},'' in \emph{Handbook on ontologies}, 2004, pp. 51--65.

\bibitem{baader2017introduction}
F.~Baader, I.~Horrocks, C.~Lutz, and U.~Sattler, \emph{Introduction to description logic}.\hskip 1em plus 0.5em minus 0.4em\relax Cambridge University Press, 2017.

\bibitem{gene2019gene}
G.~O. Consortium, ``The gene ontology resource: 20 years and still going strong,'' \emph{Nucleic acids research}, vol.~47, no.~D1, pp. D330--D338, 2019.

\bibitem{dooley2018foodon}
D.~M. Dooley, E.~J. Griffiths, G.~S. Gosal, P.~L. Buttigieg, R.~Hoehndorf, M.~C. Lange, L.~M. Schriml, F.~S. Brinkman, and W.~W. Hsiao, ``Foodon: a harmonized food ontology to increase global food traceability, quality control and data integration,'' \emph{npj Science of Food}, vol.~2, no.~1, p.~23, 2018.

\bibitem{auer2007dbpedia}
S.~Auer, C.~Bizer, G.~Kobilarov, J.~Lehmann, R.~Cyganiak, and Z.~Ives, ``{DBpedia}: A nucleus for a web of open data,'' in \emph{ISWC}, 2007, pp. 722--735.

\bibitem{garcia2011analysis}
F.~J. Garc{\'\i}a-Pe{\~n}alvo, J.~Garc{\'\i}a, R.~Ther{\'o}n~S{\'a}nchez \emph{et~al.}, ``Analysis of the {OWL} ontologies: A survey,'' 2011.

\bibitem{mikolov2013efficient}
T.~Mikolov, K.~Chen, G.~Corrado, and J.~Dean, ``Efficient estimation of word representations in vector space,'' \emph{arXiv preprint arXiv:1301.3781}, 2013.

\bibitem{bordes2013translating}
A.~Bordes, N.~Usunier, A.~Garcia-Duran, J.~Weston, and O.~Yakhnenko, ``Translating embeddings for modeling multi-relational data,'' \emph{Advances in Neural Information Processing Systems}, vol.~26, 2013.

\bibitem{ristoski2016rdf2vec}
P.~Ristoski and H.~Paulheim, ``{RDF2Vec}: {RDF} graph embeddings for data mining,'' in \emph{ISWC}, 2016, pp. 498--514.

\bibitem{wang2017knowledge}
Q.~Wang, Z.~Mao, B.~Wang, and L.~Guo, ``Knowledge graph embedding: A survey of approaches and applications,'' \emph{IEEE Transactions on Knowledge and Data Engineering}, vol.~29, no.~12, pp. 2724--2743, 2017.

\bibitem{chen2020review}
X.~Chen, S.~Jia, and Y.~Xiang, ``A review: Knowledge reasoning over knowledge graph,'' \emph{Expert Systems with Applications}, vol. 141, p. 112948, 2020.

\bibitem{nickel2017poincare}
M.~Nickel and D.~Kiela, ``Poincar{\'e} embeddings for learning hierarchical representations,'' \emph{NeurIPS}, vol.~30, 2017.

\bibitem{vilnis2018probabilistic}
L.~Vilnis, X.~Li, S.~Murty, and A.~McCallum, ``Probabilistic embedding of knowledge graphs with box lattice measures,'' in \emph{ACL}, 2018, pp. 263--272.

\bibitem{miller1995wordnet}
G.~A. Miller, ``Wordnet: a lexical database for english,'' \emph{Communications of the ACM}, vol.~38, no.~11, pp. 39--41, 1995.

\bibitem{diaz2018embeds}
G.~I. Diaz, A.~Fokoue, M.~Sadoghi \emph{et~al.}, ``Embeds: Scalable, ontology-aware graph embeddings.'' in \emph{EDBT}, 2018, pp. 433--436.

\bibitem{kulmanov2019embeddings}
M.~Kulmanov, W.~Liu-Wei, Y.~Yan, and R.~Hoehndorf, ``{EL} embeddings: Geometric construction of models for the description logic {EL++},'' in \emph{IJCAI}, 2019, pp. 6103--6109.

\bibitem{garg2019quantum}
D.~Garg, S.~Ikbal, S.~K. Srivastava, H.~Vishwakarma, H.~Karanam, and L.~V. Subramaniam, ``Quantum embedding of knowledge for reasoning,'' in \emph{NeurIPS}, 2019.

\bibitem{smaili2019opa2vec}
F.~Z. Smaili, X.~Gao, and R.~Hoehndorf, ``{OPA2Vec}: combining formal and informal content of biomedical ontologies to improve similarity-based prediction,'' \emph{Bioinformatics}, vol.~35, no.~12, pp. 2133--2140, 2019.

\bibitem{chen2021owl2vec}
J.~Chen, P.~Hu, E.~Jimenez-Ruiz, O.~M. Holter, D.~Antonyrajah, and I.~Horrocks, ``Owl2vec*: Embedding of owl ontologies,'' \emph{Machine Learning}, vol. 110, no.~7, pp. 1813--1845, 2021.

\bibitem{geng2021ontozsl}
Y.~Geng, J.~Chen, Z.~Chen, J.~Z. Pan, Z.~Ye, Z.~Yuan, Y.~Jia, and H.~Chen, ``{OntoZSL}: Ontology-enhanced zero-shot learning,'' in \emph{The Web Conference}, 2021, pp. 3325--3336.

\bibitem{kulmanov2022deepgozero}
M.~Kulmanov and R.~Hoehndorf, ``{DeepGOZero}: improving protein function prediction from sequence and zero-shot learning based on ontology axioms,'' \emph{Bioinformatics}, vol.~38, no. Supplement\_1, pp. i238--i245, 2022.

\bibitem{stoilos2005fuzzy}
G.~Stoilos, G.~B. Stamou, V.~Tzouvaras, J.~Z. Pan, and I.~Horrocks, ``Fuzzy owl: Uncertainty and the semantic web.'' in \emph{OWLED}, vol.~5, 2005, pp. 11--12.

\bibitem{maedche2001ontology}
A.~Maedche and S.~Staab, ``Ontology learning for the semantic web,'' \emph{IEEE Intelligent systems}, vol.~16, no.~2, pp. 72--79, 2001.

\bibitem{lehmann2009dl}
J.~Lehmann, ``Dl-learner: learning concepts in description logics,'' \emph{The Journal of Machine Learning Research}, vol.~10, pp. 2639--2642, 2009.

\bibitem{badreddine2022logic}
S.~Badreddine, A.~d. Garcez, L.~Serafini, and M.~Spranger, ``Logic tensor networks,'' \emph{Artificial Intelligence}, vol. 303, p. 103649, 2022.

\bibitem{zhang2022knowledge}
W.~Zhang, J.~Chen, J.~Li, Z.~Xu, J.~Z. Pan, and H.~Chen, ``Knowledge graph reasoning with logics and embeddings: survey and perspective,'' \emph{arXiv preprint arXiv:2202.07412}, 2022.

\bibitem{alam2023towards}
M.~Alam, F.~van Harmelen, and M.~Acosta, ``Towards semantically enriched embeddings for knowledge graph completion,'' \emph{arXiv:2308.00081}, 2023.

\bibitem{xiong2023geometric}
B.~Xiong, M.~Nayyeri, M.~Jin, Y.~He, M.~Cochez, S.~Pan, and S.~Staab, ``Geometric relational embeddings: A survey,'' \emph{arXiv:2304.11949}, 2023.

\bibitem{kulmanov2021semantic}
M.~Kulmanov, F.~Z. Smaili, X.~Gao, and R.~Hoehndorf, ``Semantic similarity and machine learning with ontologies,'' \emph{Briefings in Bioinformatics}, vol.~22, no.~4, p. bbaa199, 2021.

\bibitem{smaili2018onto2vec}
F.~Z. Smaili, X.~Gao, and R.~Hoehndorf, ``{Onto2Vec}: joint vector-based representation of biological entities and their ontology-based annotations,'' \emph{Bioinformatics}, vol.~34, no.~13, pp. i52--i60, 2018.

\bibitem{mowl}
F.~Zhapa-Camacho, M.~Kulmanov, and R.~Hoehndorf, ``{mOWL: Python library for machine learning with biomedical ontologies},'' \emph{Bioinformatics}, 12 2022.

\bibitem{hogan2021knowledge}
A.~Hogan, E.~Blomqvist, M.~Cochez, C.~d’Amato, G.~D. Melo, C.~Gutierrez, S.~Kirrane, J.~E.~L. Gayo, R.~Navigli, S.~Neumaier \emph{et~al.}, ``Knowledge graphs,'' \emph{ACM Computing Surveys}, vol.~54, no.~4, pp. 1--37, 2021.

\bibitem{baader2005pushing}
F.~Baader, S.~Brandt, and C.~Lutz, ``Pushing the el envelope,'' in \emph{IJCAI}, 2005, pp. 364--369.

\bibitem{donnelly2006snomed}
K.~Donnelly \emph{et~al.}, ``{SNOMED-CT}: The advanced terminology and coding system for ehealth,'' \emph{Studies in Health Technology and Informatics}, vol. 121, p. 279, 2006.

\bibitem{pennington2014glove}
J.~Pennington, R.~Socher, and C.~D. Manning, ``Glove: Global vectors for word representation,'' in \emph{EMNLP}, 2014, pp. 1532--1543.

\bibitem{vaswani2017attention}
A.~Vaswani, N.~Shazeer, N.~Parmar, J.~Uszkoreit, L.~Jones, A.~N. Gomez, {\L}.~Kaiser, and I.~Polosukhin, ``Attention is all you need,'' \emph{NeurIPS}, vol.~30, 2017.

\bibitem{kenton2019bert}
J.~D. M.-W.~C. Kenton and L.~K. Toutanova, ``{BERT}: Pre-training of deep bidirectional transformers for language understanding,'' in \emph{NAACL}, 2019, pp. 4171--4186.

\bibitem{asgari2015continuous}
E.~Asgari and M.~R. Mofrad, ``Continuous distributed representation of biological sequences for deep proteomics and genomics,'' \emph{PloS one}, vol.~10, no.~11, p. e0141287, 2015.

\bibitem{grover2016node2vec}
A.~Grover and J.~Leskovec, ``{Node2Vec}: Scalable feature learning for networks,'' in \emph{ACM SIGKDD}, 2016, pp. 855--864.

\bibitem{yang2015embedding}
B.~Yang, S.~W.-t. Yih, X.~He, J.~Gao, and L.~Deng, ``Embedding entities and relations for learning and inference in knowledge bases,'' in \emph{ICLR}, 2015.

\bibitem{dettmers2018convolutional}
T.~Dettmers, P.~Minervini, P.~Stenetorp, and S.~Riedel, ``Convolutional 2d knowledge graph embeddings,'' in \emph{AAAI}, vol.~32, no.~1, 2018.

\bibitem{gesese2021survey}
G.~A. Gesese, R.~Biswas, M.~Alam, and H.~Sack, ``A survey on knowledge graph embeddings with literals: Which model links better literal-ly?'' \emph{Semantic Web}, vol.~12, no.~4, pp. 617--647, 2021.

\bibitem{pan2023large}
J.~Z. Pan, S.~Razniewski, J.-C. Kalo, S.~Singhania, J.~Chen, S.~Dietze, H.~Jabeen, J.~Omeliyanenko, W.~Zhang, M.~Lissandrini \emph{et~al.}, ``Large language models and knowledge graphs: Opportunities and challenges,'' \emph{Transactions on Graph Data and Knowledge}, pp. 1--38, 2023.

\bibitem{bengio2013representation}
Y.~Bengio, A.~Courville, and P.~Vincent, ``Representation learning: A review and new perspectives,'' \emph{IEEE Transactions on Pattern Analysis and Machine Intelligence}, vol.~35, no.~8, pp. 1798--1828, 2013.

\bibitem{dlhandbook}
\emph{The Description Logic Handbook: Theory, Implementation and Applications}.\hskip 1em plus 0.5em minus 0.4em\relax Cambridge University Press, Aug. 2007.

\bibitem{linardatos2020explainable}
P.~Linardatos, V.~Papastefanopoulos, and S.~Kotsiantis, ``Explainable ai: A review of machine learning interpretability methods,'' \emph{Entropy}, vol.~23, no.~1, p.~18, 2020.

\bibitem{doshi2017towards}
F.~Doshi-Velez and B.~Kim, ``Towards a rigorous science of interpretable machine learning,'' \emph{arXiv:1702.08608}, 2017.

\bibitem{peng2022description}
X.~Peng, Z.~Tang, M.~Kulmanov, K.~Niu, and R.~Hoehndorf, ``Description logic {EL++} embeddings with intersectional closure,'' \emph{arXiv:2202.14018}, 2022.

\bibitem{jackermeier2024dual}
M.~Jackermeier, J.~Chen, and I.~Horrocks, ``Dual box embeddings for the description logic el++,'' in \emph{The Web Conference}, 2024.

\bibitem{hao2023ontology}
Z.~Hao, W.~Mayer, J.~Xia, G.~Li, L.~Qin, and Z.~Feng, ``Ontology alignment with semantic and structural embeddings,'' \emph{Journal of Web Semantics}, vol.~78, p. 100798, 2023.

\bibitem{vendrov2015order}
I.~Vendrov, R.~Kiros, S.~Fidler, and R.~Urtasun, ``Order-embeddings of images and language,'' in \emph{ICML}, 2016.

\bibitem{athiwaratkun2018hierarchical}
B.~Athiwaratkun and A.~G. Wilson, ``Hierarchical density order embeddings,'' in \emph{ICLR}, 2018.

\bibitem{li2018smoothing}
X.~Li, L.~Vilnis, D.~Zhang, M.~Boratko, and A.~McCallum, ``Smoothing the geometry of probabilistic box embeddings,'' in \emph{ICLR}, 2018.

\bibitem{patel2020representing}
D.~Patel and S.~Sankar, ``Representing joint hierarchies with box embeddings,'' \emph{Automated Knowledge Base Construction}, 2020.

\bibitem{dasgupta2020improving}
S.~Dasgupta, M.~Boratko, D.~Zhang, L.~Vilnis, X.~Li, and A.~McCallum, ``Improving local identifiability in probabilistic box embeddings,'' \emph{NeurIPS}, vol.~33, pp. 182--192, 2020.

\bibitem{ganea2018hyperbolic}
O.~Ganea, G.~B{\'e}cigneul, and T.~Hofmann, ``Hyperbolic entailment cones for learning hierarchical embeddings,'' in \emph{ICML}, 2018, pp. 1646--1655.

\bibitem{lu2019learning}
Q.~Lu, N.~De~Silva, S.~Kafle, J.~Cao, D.~Dou, T.~H. Nguyen, P.~Sen, B.~Hailpern, B.~Reinwald, and Y.~Li, ``Learning electronic health records through hyperbolic embedding of medical ontologies,'' in \emph{ACM-BCB}, 2019, pp. 338--346.

\bibitem{li2021hypon}
Z.~Li and S.~Wang, ``{HYPON}: embedding biomedical ontology with entity sets,'' in \emph{ACM-BCB}, 2021, pp. 1--7.

\bibitem{pan2021hyperbolic}
Z.~Pan and P.~Wang, ``Hyperbolic hierarchy-aware knowledge graph embedding for link prediction,'' in \emph{Findings of the EMNLP}, 2021, pp. 2941--2948.

\bibitem{gutierrez2018from}
V.~Gutierrez~Basulto and S.~Schockaert, ``From knowledge graph embedding to ontology embedding? an analysis of the compatibility between vector space representations and rules,'' in \emph{KR}, 2018.

\bibitem{ozccep2020cone}
{\"O}.~L. {\"O}z{\c{c}}ep, M.~Leemhuis, and D.~Wolter, ``Cone semantics for logics with negation,'' in \emph{IJCAI}, 2020, pp. 1820--1826.

\bibitem{mondal2021emel++}
S.~Mondal, S.~Bhatia, and R.~Mutharaju, ``{EmEL++}: Embeddings for $\varepsilon$l++ description logic,'' in \emph{AAAI Spring Symposium on Combining Machine Learning and Knowledge Engineering}, 2021.

\bibitem{xiong2022faithful}
B.~Xiong, N.~Potyka, T.-K. Tran, M.~Nayyeri, and S.~Staab, ``Faithful embeddings for el++ knowledge bases,'' in \emph{ISWC}, 2022, pp. 22--38.

\bibitem{ozcep2023embedding}
{\"O}.~L. {\"O}zcep, M.~Leemhuis, and D.~Wolter, ``Embedding ontologies in the description logic alc by axis-aligned cones,'' \emph{Journal of Artificial Intelligence Research}, vol.~78, pp. 217--267, 2023.

\bibitem{lacerda2023strong}
V.~Lacerda, A.~Ozaki, and R.~Guimar{\~a}es, ``Strong faithfulness for elh ontology embeddings,'' \emph{arXiv:2310.02198}, 2023.

\bibitem{zhapa2023cate}
F.~Zhapa-Camacho and R.~Hoehndorf, ``Lattice-preserving $\mathcal{ALC}$ ontology embeddings,'' in \emph{Neural-Symbolic Learning and Reasoning}, T.~R. Besold, A.~d'Avila Garcez, E.~Jimenez-Ruiz, R.~Confalonieri, P.~Madhyastha, and B.~Wagner, Eds., 2024, pp. 355--369.

\bibitem{lv2018differentiating}
X.~Lv, L.~Hou, J.~Li, and Z.~Liu, ``Differentiating concepts and instances for knowledge graph embedding,'' in \emph{EMNLP}, 2018, pp. 1971--1979.

\bibitem{hao2019universal}
J.~Hao, M.~Chen, W.~Yu, Y.~Sun, and W.~Wang, ``Universal representation learning of knowledge bases by jointly embedding instances and ontological concepts,'' in \emph{ACM SIGKDD}, 2019, pp. 1709--1719.

\bibitem{xiang2021ontoea}
Y.~Xiang, Z.~Zhang, J.~Chen, X.~Chen, Z.~Lin, and Y.~Zheng, ``{OntoEA}: Ontology-guided entity alignment via joint knowledge graph embedding,'' in \emph{Findings of the ACL}, 2021, pp. 1117--1128.

\bibitem{yu2023geometry}
J.~Yu, C.~Zhang, Z.~Hu, Y.~Ji, D.~Fu, and X.~Wang, ``Geometry-based anisotropy representation learning of concepts for knowledge graph embedding,'' \emph{Applied Intelligence}, vol.~53, no.~17, pp. 19\,940--19\,961, 2023.

\bibitem{huang2023concept2box}
Z.~Huang, D.~Wang, B.~Huang, C.~Zhang, J.~Shang, Y.~Liang, Z.~Wang, X.~Li, C.~Faloutsos, Y.~Sun \emph{et~al.}, ``{Concept2Box}: Joint geometric embeddings for learning two-view knowledge graphs,'' \emph{Findings of the ACL}, 2023.

\bibitem{wang2024embedding}
K.~Wang, G.~Qi, J.~Chen, and T.~Wu, ``Embedding ontologies via incoprorating extensional and intensional knowledge,'' \emph{arXiv:2402.01677}, 2024.

\bibitem{alshahrani2017neuro}
M.~Alshahrani, M.~A. Khan, O.~Maddouri, A.~R. Kinjo, N.~Queralt-Rosinach, and R.~Hoehndorf, ``Neuro-symbolic representation learning on biological knowledge graphs,'' \emph{Bioinformatics}, vol.~33, no.~17, pp. 2723--2730, 2017.

\bibitem{holter2019embedding}
O.~M. Holter, E.~B. Myklebust, J.~Chen, and E.~Jim{\'e}nez-Ruiz, ``Embedding {OWL} ontologies with {OWL2Vec},'' in \emph{CEUR Workshop Proceedings}, vol. 2456, 2019, pp. 33--36.

\bibitem{xiang2015ersom}
C.~Xiang, T.~Jiang, B.~Chang, and Z.~Sui, ``{ERSOM}: A structural ontology matching approach using automatically learned entity representation,'' in \emph{EMNLP}, 2015, pp. 2419--2429.

\bibitem{jayawardana2017deriving}
V.~Jayawardana, D.~Lakmal, N.~de~Silva, A.~S. Perera, K.~Sugathadasa, and B.~Ayesha, ``Deriving a representative vector for ontology classes with instance word vector embeddings,'' in \emph{INTECH}, 2017, pp. 79--84.

\bibitem{kolyvakis2018deepalignment}
P.~Kolyvakis, A.~Kalousis, and D.~Kiritsis, ``{DeepAlignment}: Unsupervised ontology matching with refined word vectors,'' in \emph{NAACL}, 2018, pp. 787--798.

\bibitem{dong2019imposing}
T.~Dong, C.~Bauckhage, H.~Jin, J.~Li, O.~Cremers, D.~Speicher, A.~B. Cremers, and J.~Zimmermann, ``Imposing category trees onto word-embeddings using a geometric construction,'' in \emph{ICLR}, 2019.

\bibitem{liu2020concept}
H.~Liu, Y.~Perl, and J.~Geller, ``Concept placement using {BERT} trained by transforming and summarizing biomedical ontology structure,'' \emph{Journal of Biomedical Informatics}, vol. 112, p. 103607, 2020.

\bibitem{nguyen2021biomedical}
V.~Nguyen, H.~Y. Yip, and O.~Bodenreider, ``Biomedical vocabulary alignment at scale in the {UMLS} metathesaurus,'' in \emph{The Web Conference}, 2021, pp. 2672--2683.

\bibitem{liu2021self}
F.~Liu, E.~Shareghi, Z.~Meng, M.~Basaldella, and N.~Collier, ``Self-alignment pretraining for biomedical entity representations,'' in \emph{NAACL}, 2021, pp. 4228--4238.

\bibitem{zhang2022ontoprotein}
N.~Zhang, Z.~Bi, X.~Liang, S.~Cheng, H.~Hong, S.~Deng, Q.~Zhang, J.~Lian, and H.~Chen, ``{OntoProtein}: Protein pretraining with gene ontology embedding,'' in \emph{ICLR}, 2022.

\bibitem{chen2023contextual}
J.~Chen, Y.~He, Y.~Geng, E.~Jim{\'e}nez-Ruiz, H.~Dong, and I.~Horrocks, ``Contextual semantic embeddings for ontology subsumption prediction,'' \emph{World Wide Web}, vol.~26, no.~5, pp. 2569--2591, 2023.

\bibitem{gosselin2023sorbet}
F.~Gosselin and A.~Zouaq, ``{SORBET}: A siamese network for ontology embeddings using a distance-based regression loss and bert,'' in \emph{ISWC}, 2023, pp. 561--578.

\bibitem{he2024language}
Y.~He, Z.~Yuan, J.~Chen, and I.~Horrocks, ``Language models as hierarchy encoders,'' \emph{arXiv:2401.11374}, 2024.

\bibitem{ma2021hyperexpan}
M.~D. Ma, M.~Chen, T.-L. Wu, and N.~Peng, ``{HyperExpan}: Taxonomy expansion with hyperbolic representation learning,'' in \emph{Findings of the EMNLP}, 2021, pp. 4182--4194.

\bibitem{hao2021medto}
J.~Hao, C.~Lei, V.~Efthymiou, A.~Quamar, F.~{\"O}zcan, Y.~Sun, and W.~Wang, ``Medto: Medical data to ontology matching using hybrid graph neural networks,'' in \emph{ACM SIGKDD}, 2021, pp. 2946--2954.

\bibitem{nickel2018learning}
M.~Nickel and D.~Kiela, ``Learning continuous hierarchies in the lorentz model of hyperbolic geometry,'' in \emph{ICML}, 2018, pp. 3779--3788.

\bibitem{chami2019hyperbolic}
I.~Chami, Z.~Ying, C.~R{\'e}, and J.~Leskovec, ``Hyperbolic graph convolutional neural networks,'' \emph{NeurIPS}, vol.~32, 2019.

\bibitem{tang2022falcon}
Z.~Tang, T.~Hinnerichs, X.~Peng, X.~Zhang, and R.~Hoehndorf, ``Falcon: Sound and complete neural semantic entailment over alc ontologies,'' 2022.

\bibitem{abboud2020boxe}
R.~Abboud, I.~Ceylan, T.~Lukasiewicz, and T.~Salvatori, ``{BoxE}: A box embedding model for knowledge base completion,'' \emph{NeurIPS}, vol.~33, pp. 9649--9661, 2020.

\bibitem{he2022bertmap}
Y.~He, J.~Chen, D.~Antonyrajah, and I.~Horrocks, ``{BERTMap}: a bert-based ontology alignment system,'' in \emph{AAAI}, vol.~36, no.~5, 2022, pp. 5684--5691.

\bibitem{zhang2019iteratively}
W.~Zhang, B.~Paudel, L.~Wang, J.~Chen, H.~Zhu, W.~Zhang, A.~Bernstein, and H.~Chen, ``Iteratively learning embeddings and rules for knowledge graph reasoning,'' in \emph{The World Wide Web Conference}, 2019, pp. 2366--2377.

\bibitem{xie2016representation}
R.~Xie, Z.~Liu, M.~Sun \emph{et~al.}, ``Representation learning of knowledge graphs with hierarchical types.'' in \emph{IJCAI}, vol. 2016, 2016, pp. 2965--2971.

\bibitem{otero2015ontology}
L.~Otero-Cerdeira, F.~J. Rodr{\'\i}guez-Mart{\'\i}nez, and A.~G{\'o}mez-Rodr{\'\i}guez, ``Ontology matching: A literature review,'' \emph{Expert Systems with Applications}, vol.~42, no.~2, pp. 949--971, 2015.

\bibitem{he2022machine}
Y.~He, J.~Chen, H.~Dong, E.~Jim{\'e}nez-Ruiz, A.~Hadian, and I.~Horrocks, ``Machine learning-friendly biomedical datasets for equivalence and subsumption ontology matching,'' in \emph{ISWC}, 2022, pp. 575--591.

\bibitem{chen2021augmenting}
J.~Chen, E.~Jim{\'e}nez-Ruiz, I.~Horrocks, D.~Antonyrajah, A.~Hadian, and J.~Lee, ``Augmenting ontology alignment by semantic embedding and distant supervision,'' in \emph{ESWC}, 2021, pp. 392--408.

\bibitem{ritchie2021ontology}
A.~Ritchie, J.~Chen, L.~J. Castro, D.~Rebholz-Schuhmann, and E.~Jim{\'e}nez-Ruiz, ``Ontology clustering with owl2vec,'' in \emph{CEUR Workshop Proceedings}, vol. 2918, 2021, pp. 54--61.

\bibitem{chen2023zero}
J.~Chen, Y.~Geng, Z.~Chen, J.~Z. Pan, Y.~He, W.~Zhang, I.~Horrocks, and H.~Chen, ``Zero-shot and few-shot learning with knowledge graphs: A comprehensive survey,'' \emph{Proceedings of the IEEE}, 2023.

\bibitem{chen2021knowledge}
J.~Chen, Y.~Geng, Z.~Chen, I.~Horrocks, J.~Z. Pan, and H.~Chen, ``Knowledge-aware zero-shot learning: Survey and perspective,'' in \emph{IJCAI}, 2021.

\bibitem{xian2017zero}
Y.~Xian, B.~Schiele, and Z.~Akata, ``Zero-shot learning-the good, the bad and the ugly,'' in \emph{CVPR}, 2017, pp. 4582--4591.

\bibitem{chen2020ontology}
J.~Chen, F.~L{\'e}cu{\'e}, Y.~Geng, J.~Z. Pan, and H.~Chen, ``Ontology-guided semantic composition for zero-shot learning,'' in \emph{KR}, vol.~17, no.~1, 2020, pp. 850--854.

\bibitem{Robinson_2008}
P.~N. Robinson, S.~Köhler, S.~Bauer, D.~Seelow, D.~Horn, and S.~Mundlos, ``The human phenotype ontology: A tool for annotating and analyzing human hereditary disease,'' \emph{The American Journal of Human Genetics}, vol.~83, no.~5, p. 610–615, Nov. 2008.

\bibitem{choi2017gram}
E.~Choi, M.~T. Bahadori, L.~Song, W.~F. Stewart, and J.~Sun, ``Gram: graph-based attention model for healthcare representation learning,'' in \emph{ACM SIGKDD}, 2017, pp. 787--795.

\bibitem{peng2021sequential}
X.~Peng, G.~Long, T.~Shen, S.~Wang, and J.~Jiang, ``Sequential diagnosis prediction with transformer and ontological representation,'' in \emph{ICDM}, 2021, pp. 489--498.

\bibitem{niu2022fusion}
K.~Niu, Y.~Lu, X.~Peng, and J.~Zeng, ``Fusion of sequential visits and medical ontology for mortality prediction,'' \emph{Journal of Biomedical Informatics}, vol. 127, p. 104012, 2022.

\bibitem{cheong2023adaptive}
C.~W. Cheong, K.~Yin, W.~K. Cheung, B.~C. Fung, and J.~Poon, ``Adaptive integration of categorical and multi-relational ontologies with ehr data for medical concept embedding,'' \emph{ACM Transactions on Intelligent Systems and Technology}, vol.~14, no.~6, pp. 1--20, 2023.

\bibitem{zhang2020hierarchical}
M.~Zhang, C.~R. King, M.~Avidan, and Y.~Chen, ``Hierarchical attention propagation for healthcare representation learning,'' in \emph{ACM SIGKDD}, 2020, pp. 249--256.

\bibitem{song2019medical}
L.~Song, C.~W. Cheong, K.~Yin, W.~K. Cheung, B.~C. Fung, and J.~Poon, ``Medical concept embedding with multiple ontological representations.'' in \emph{IJCAI}, vol.~19, 2019, pp. 4613--4619.

\bibitem{ma2018kame}
F.~Ma, Q.~You, H.~Xiao, R.~Chitta, J.~Zhou, and J.~Gao, ``Kame: Knowledge-based attention model for diagnosis prediction in healthcare,'' in \emph{CIKM}, 2018, pp. 743--752.

\bibitem{yao2023ontology}
Z.~Yao, B.~Liu, F.~Wang, D.~Sow, and Y.~Li, ``Ontology-aware prescription recommendation in treatment pathways using multi-evidence healthcare data,'' \emph{ACM Transactions on Information Systems}, vol.~41, no.~4, pp. 1--29, 2023.

\bibitem{Shen_2019}
F.~Shen, S.~Peng, Y.~Fan, A.~Wen, S.~Liu, Y.~Wang, L.~Wang, and H.~Liu, ``Hpo2vec+: Leveraging heterogeneous knowledge resources to enrich node embeddings for the human phenotype ontology,'' \emph{Journal of Biomedical Informatics}, vol.~96, p. 103246, 2019.

\bibitem{Nunes_2023}
S.~Nunes, R.~T. Sousa, and C.~Pesquita, ``Multi-domain knowledge graph embeddings for gene-disease association prediction,'' \emph{Journal of Biomedical Semantics}, vol.~14, no.~1, Aug. 2023.

\bibitem{embedpvp}
A.~Althagafi, F.~Zhapa-Camacho, and R.~Hoehndorf, ``{Prioritizing genomic variants through neuro-symbolic, knowledge-enhanced learning},'' \emph{Bioinformatics}, p. btae301, 05 2024.

\bibitem{agarwal2019snomed2vec}
K.~Agarwal, T.~Eftimov, R.~Addanki, S.~Choudhury, S.~Tamang, and R.~Rallo, ``Snomed2vec: Random walk and poincar$\backslash$'e embeddings of a clinical knowledge base for healthcare analytics,'' \emph{arXiv:1907.08650}, 2019.

\bibitem{Vilela_2022}
J.~Vilela, M.~Asif, A.~R. Marques, J.~X. Santos, C.~Rasga, A.~Vicente, and H.~Martiniano, ``Biomedical knowledge graph embeddings for personalized medicine: Predicting disease‐gene associations,'' \emph{Expert Systems}, vol.~40, no.~5, Nov. 2022.

\bibitem{Wang_2023}
Y.~Wang, P.~Wegner, D.~Domingo-Fernández, and A.~Tom~Kodamullil, ``Multi-ontology embeddings approach on human-aligned multi-ontologies representation for gene-disease associations prediction,'' \emph{Heliyon}, vol.~9, no.~11, p. e21502, Nov. 2023.

\bibitem{zhao2022}
C.~Zhao, T.~Liu, and Z.~Wang, ``{PANDA2: protein function prediction using graph neural networks},'' \emph{NAR Genomics and Bioinformatics}, vol.~4, no.~1, p. lqac004, 02 2022.

\bibitem{qiu2022isoform}
S.~Qiu, G.~Yu, X.~Lu, C.~Domeniconi, and M.~Guo, ``{Isoform function prediction by Gene Ontology embedding},'' \emph{Bioinformatics}, vol.~38, no.~19, pp. 4581--4588, 08 2022.

\bibitem{Kulmanov_2024}
M.~Kulmanov, F.~J. Guzmán-Vega, P.~Duek~Roggli, L.~Lane, S.~T. Arold, and R.~Hoehndorf, ``Protein function prediction as approximate semantic entailment,'' \emph{Nature Machine Intelligence}, vol.~6, no.~2, p. 220–228, Feb. 2024.

\bibitem{li2024}
W.~Li, B.~Wang, J.~Dai, Y.~Kou, X.~Chen, Y.~Pan, S.~Hu, and Z.~Z. Xu, ``{Partial order relation–based gene ontology embedding improves protein function prediction},'' \emph{Briefings in Bioinformatics}, vol.~25, no.~2, p. bbae077, 03 2024.

\bibitem{esm2}
Z.~Lin, H.~Akin, R.~Rao, B.~Hie, Z.~Zhu, W.~Lu, N.~Smetanin, R.~Verkuil, O.~Kabeli, Y.~Shmueli, A.~dos Santos~Costa, M.~Fazel-Zarandi, T.~Sercu, S.~Candido, and A.~Rives, ``Evolutionary-scale prediction of atomic-level protein structure with a language model,'' \emph{Science}, vol. 379, no. 6637, pp. 1123--1130, 2023.

\bibitem{prottrans}
A.~Elnaggar, M.~Heinzinger, C.~Dallago, G.~Rehawi, Y.~Wang, L.~Jones, T.~Gibbs, T.~Feher, C.~Angerer, M.~Steinegger, D.~Bhowmik, and B.~Rost, ``Prottrans: Toward understanding the language of life through self-supervised learning,'' \emph{IEEE Transactions on Pattern Analysis and Machine Intelligence}, vol.~44, no.~10, pp. 7112--7127, 2022.

\bibitem{salatino2019cso}
A.~A. Salatino, F.~Osborne, T.~Thanapalasingam, and E.~Motta, ``The cso classifier: Ontology-driven detection of research topics in scholarly articles,'' in \emph{TPDL}.\hskip 1em plus 0.5em minus 0.4em\relax Springer, 2019, pp. 296--311.

\bibitem{ali2019fuzzy}
F.~Ali, S.~El-Sappagh, and D.~Kwak, ``Fuzzy ontology and lstm-based text mining: a transportation network monitoring system for assisting travel,'' \emph{Sensors}, vol.~19, no.~2, p. 234, 2019.

\bibitem{sweidan2021sentence}
A.~H. Sweidan, N.~El-Bendary, and H.~Al-Feel, ``Sentence-level aspect-based sentiment analysis for classifying adverse drug reactions (adrs) using hybrid ontology-xlnet transfer learning,'' \emph{IEEE Access}, vol.~9, pp. 90\,828--90\,846, 2021.

\bibitem{deng2021ontoed}
S.~Deng, N.~Zhang, L.~Li, H.~Chen, H.~Tou, M.~Chen, F.~Huang, and H.~Chen, ``{OntoED}: Low-resource event detection with ontology embedding,'' \emph{arXiv:2105.10922}, 2021.

\bibitem{erten2021ontology}
C.~Erten and D.~Kazakov, ``Ontology graph embeddings and ilp for financial forecasting,'' in \emph{ILP}.\hskip 1em plus 0.5em minus 0.4em\relax Springer, 2021, pp. 111--124.

\bibitem{dassereto2020evaluating}
F.~Dassereto, L.~Di~Rocco, G.~Guerrini, and M.~Bertolotto, ``Evaluating the effectiveness of embeddings in representing the structure of geospatial ontologies,'' in \emph{The AGILE Conference on Geographic Information Science}.\hskip 1em plus 0.5em minus 0.4em\relax Springer, 2020, pp. 41--57.

\bibitem{wick2015geonames}
M.~Wick, B.~Vatant, and B.~Christophe, ``Geonames ontology,'' \emph{URL http://www. geonames. org/ontology}, 2015.

\bibitem{owlapi}
M.~Horridge and S.~Bechhofer, ``{The OWL API: A Java API for OWL Ontologies},'' \emph{Semant. Web}, vol.~2, no.~1, p. 11–21, jan 2011.

\bibitem{van_Bekkum_2021}
M.~van Bekkum, M.~de~Boer, F.~van Harmelen, A.~Meyer-Vitali, and A.~t. Teije, ``Modular design patterns for hybrid learning and reasoning systems: a taxonomy, patterns and use cases,'' \emph{Applied Intelligence}, vol.~51, no.~9, p. 6528–6546, Jun. 2021.

\bibitem{chang2024survey}
Y.~Chang, X.~Wang, J.~Wang, Y.~Wu, L.~Yang, K.~Zhu, H.~Chen, X.~Yi, C.~Wang, Y.~Wang \emph{et~al.}, ``A survey on evaluation of large language models,'' \emph{ACM Transactions on Intelligent Systems and Technology}, vol.~15, no.~3, pp. 1--45, 2024.

\bibitem{touvron2023llama}
H.~Touvron, T.~Lavril, G.~Izacard, X.~Martinet, M.-A. Lachaux, T.~Lacroix, B.~Rozi{\`e}re, N.~Goyal, E.~Hambro, F.~Azhar \emph{et~al.}, ``Llama: Open and efficient foundation language models,'' \emph{arXiv:2302.13971}, 2023.

\bibitem{pan2024unifying}
S.~Pan, L.~Luo, Y.~Wang, C.~Chen, J.~Wang, and X.~Wu, ``Unifying large language models and knowledge graphs: A roadmap,'' \emph{IEEE Transactions on Knowledge and Data Engineering}, 2024.

\bibitem{lewis2020retrieval}
P.~Lewis, E.~Perez, A.~Piktus, F.~Petroni, V.~Karpukhin, N.~Goyal, H.~K{\"u}ttler, M.~Lewis, W.-t. Yih, T.~Rockt{\"a}schel \emph{et~al.}, ``Retrieval-augmented generation for knowledge-intensive nlp tasks,'' \emph{Advances in Neural Information Processing Systems}, vol.~33, pp. 9459--9474, 2020.

\bibitem{hitzler2020neural}
P.~Hitzler, F.~Bianchi, M.~Ebrahimi, and M.~K. Sarker, ``Neural-symbolic integration and the semantic web,'' \emph{Semantic Web}, vol.~11, no.~1, pp. 3--11, 2020.

\bibitem{bian2021generative}
Y.~Bian and X.-Q. Xie, ``Generative chemistry: drug discovery with deep learning generative models,'' \emph{Journal of Molecular Modeling}, vol.~27, pp. 1--18, 2021.

\bibitem{jullien2023semeval}
M.~Jullien, M.~Valentino, H.~Frost, P.~O'Regan, D.~Landers, and A.~Freitas, ``Semeval-2023 task 7: Multi-evidence natural language inference for clinical trial data,'' \emph{arXiv:2305.02993}, 2023.

\end{thebibliography}
\begin{IEEEbiography}
[{\includegraphics[width=0.5in,height=0.75in,clip,keepaspectratio]{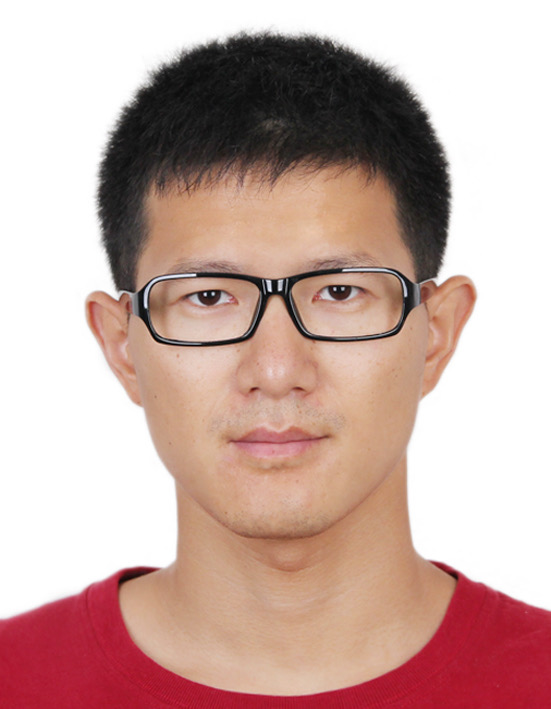}}]{Jiaoyan Chen} is a Lecturer in Department of Computer Science, University of Manchester and a part-time Senior Researcher in Department of Computer Science, University of Oxford. His main interests include KG, ontology and machine learning.
\end{IEEEbiography}
\vspace{-1cm}
\begin{IEEEbiography}
[{\includegraphics[width=0.75in,height=0.5in,clip,keepaspectratio]{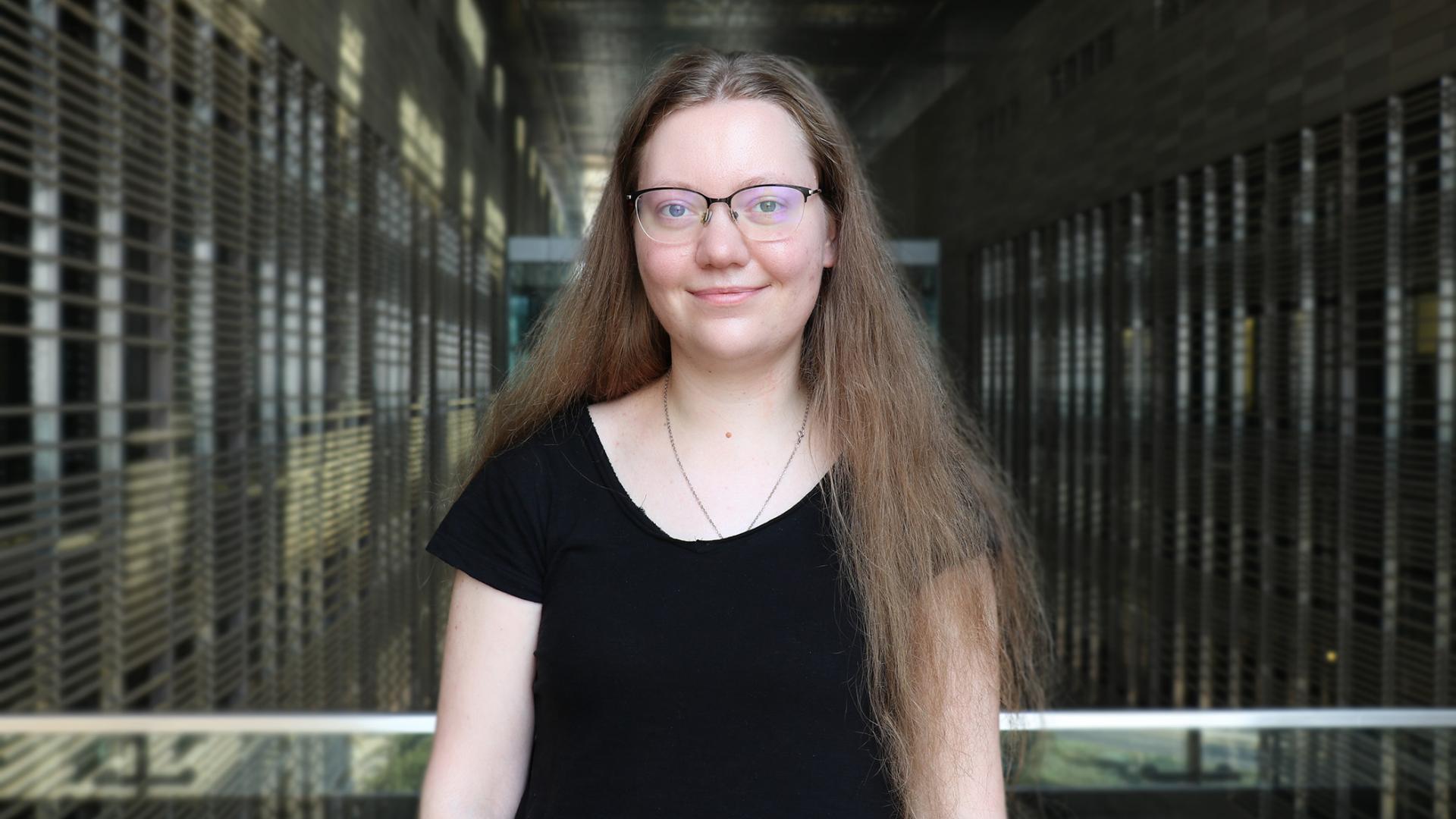}}]{Olga Mashkova} is a PhD student in Computer Science at King Abdullah University of Science and Technology. Her main interests include deep learning, bioinformatics and knowledge representation and reasoning.
\end{IEEEbiography}
\vspace{-1cm}
\begin{IEEEbiography}
[{\includegraphics[width=0.75in,height=0.5in,clip,keepaspectratio]{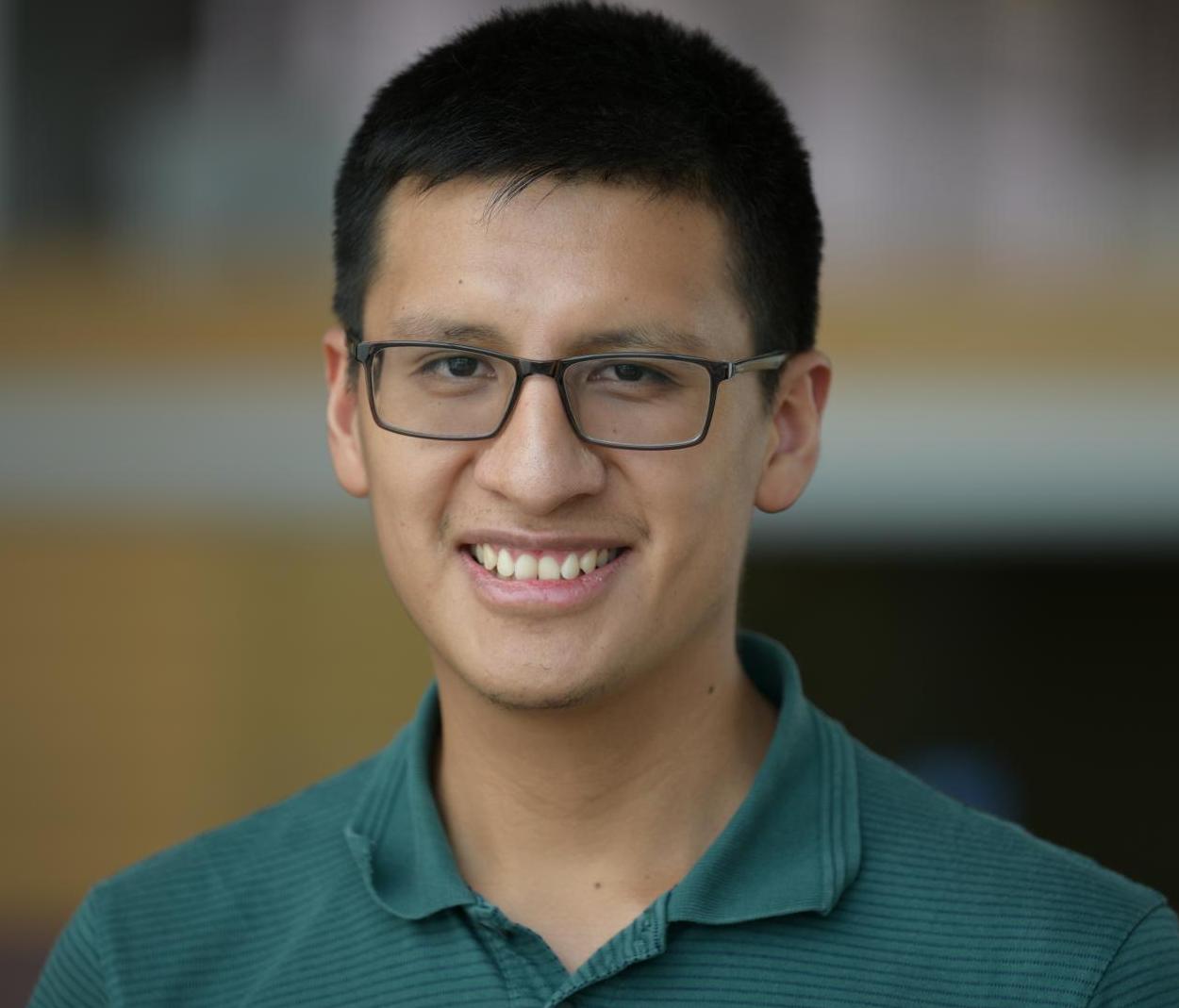}}]{Fernando Zhapa-Camacho} is a PhD student in Computer Science at King Abdullah University of Science and Technology. His main interests include knowledge representation and reasoning, machine learning and bioinformatics.
\end{IEEEbiography}
\vspace{-1cm}
\begin{IEEEbiography}
[{\includegraphics[width=0.75in,height=0.5in,clip,keepaspectratio]{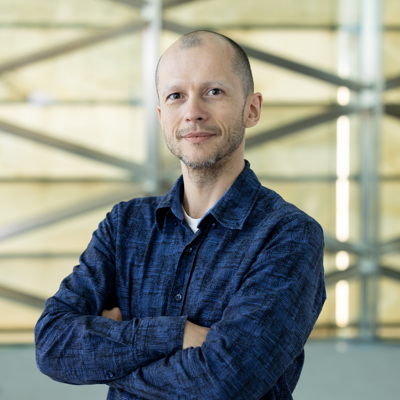}}]{Robert Hoehndorf} is an Associate Professor in Computer Science at King Abdullah University of Science and Technology. His main interests include artificial intelligence, knowledge representation, biomedical informatics and ontology.
\end{IEEEbiography}
\vspace{-1cm}
\begin{IEEEbiography}
[{\includegraphics[width=0.5in,height=0.75in,clip,keepaspectratio]{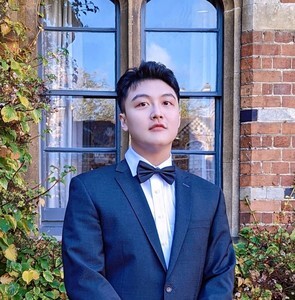}}]{Yuan He} is a Researcher in Department of Computer Science, University of Oxford. His main interests include Large Language Models, knowledge engineering and neural-symbolic integration.
\end{IEEEbiography}
\vspace{-1cm}
\begin{IEEEbiography}
[{\includegraphics[width=0.5in,height=0.75in,clip,keepaspectratio]{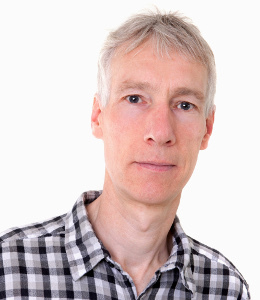}}]{Ian Horrocks} is a Professor in Computer Science, University of Oxford. His main interests include  knowledge representation, ontologies and ontology languages, description logics, and the Semantic Web.
\end{IEEEbiography}

\end{document}